\documentclass{article}                                                                           

\renewcommand\thesection{\Roman{section}.}                                         
\renewcommand\thesubsection{\thesection\Alph{subsection}.}                   
\renewcommand\thesubsubsection{\thesubsection\arabic{subsubsection}.} 

\usepackage[explicit]{titlesec}                                                                  
\titleformat{\section}{\bfseries}{\thesection}{1em}{\MakeUppercase{#1}}  
\titleformat{\subsubsection}{\itshape}{\thesubsubsection}{1em}{#1}          

\usepackage{setspace}                                                                            
\usepackage{indentfirst}                                                                          
\usepackage[margin=1.3in]{geometry}                                                     

\usepackage{lastpage}                                                                             
\usepackage[figure,table]{totalcount}                                                        

\usepackage{amsmath}                                                                            
\usepackage{multirow}                                                                             
\usepackage{graphicx}                                                                             

\usepackage[]{footmisc}                                                                          

\usepackage{caption}                                                                              
\usepackage[labelformat=simple]{subcaption}                                           
\usepackage[hidelinks]{hyperref}
\usepackage{cleveref}     

\crefname{algocf}{alg.}{algs.}
\Crefname{algocf}{Algorithm}{Algorithms}
\usepackage{paralist}
\usepackage[ruled,vlined,linesnumbered]{algorithm2e}
\usepackage[noend]{algpseudocode}

\usepackage{amssymb,amsthm}
\usepackage{epsfig}
\usepackage[mathcal]{euscript}
\usepackage{mathrsfs}
\newenvironment{myalign}{\par\nobreak\small\noindent\align}{\endalign}
\usepackage{esvect}
\usepackage{siunitx}

\usepackage{setspace}
\usepackage{color}
\usepackage{array}
\usepackage{pdflscape} 

\usepackage{float} 
\usepackage{xspace}

\makeatletter
\newcommand{\mypm}{\mathbin{\mathpalette\@mypm\relax}}
\newcommand{\@mypm}[2]{\ooalign{%
  \raisebox{.1\height}{$#1+$}\cr
  \smash{\raisebox{-.6\height}{$#1-$}}\cr}}
\makeatother

\usepackage{url}
\makeatletter
\g@addto@macro{\UrlBreaks}{\UrlOrds}
\makeatother

\usepackage{tabularx}
\usepackage{ragged2e}  
\newcolumntype{Y}{>{\RaggedRight\arraybackslash}X} 
\usepackage{booktabs}  
\usepackage[utf8]{inputenc}
\usepackage[english]{babel}

\usepackage{tikz}
\usepackage{xstring,calc}
\usetikzlibrary{calc}

\tikzset{DNA Style/.style={minimum size=0.5cm, draw=gray, line width=1pt, inner sep = 2pt}}{}

\newcounter{ColumnCounter}

\newcommand*{\PreviousNode}{}%
\newcommand*{\Sequence}[2][Mark]{%
    \def\Sequence{#2}%
    \def\PreviousNode{}%
    \foreach [count=\xi] \Label/\Color in \Sequence {%
        \IfStrEq{\Color}{}{\def\Color{white}}{}
        \edef\NodeName{#1-\arabic{ColumnCounter}}
        \IfStrEq{\PreviousNode}{}{%
            \node [DNA Style, fill=\Color, anchor=west] (\NodeName) {\Label};
            \xdef\PreviousNode{\NodeName}%
        }{
            \node [DNA Style, fill=\Color, anchor=west, xshift=-\pgflinewidth] at (\PreviousNode.east)(\NodeName) {\Label};
            \xdef\PreviousNode{\NodeName}%
        }
        \stepcounter{ColumnCounter}
    } 
}%

\usepackage[draft]{todonotes}   

\begin{document}

\title{Gnowee: A Hybrid Metaheuristic Optimization Algorithm for Constrained, Black Box, Combinatorial Mixed-Integer Engineering Design Problems}


\author{James~E.~Bevins, %
        R.\ N.\ Slaybaugh%
\thanks{J.\ Bevins is with the Department
of Nuclear Engineering, University of California, Berkeley, Berkeley,
CA, 94720 USA e-mail: (email: james.e.bevins@gmail.com).}
\thanks{R.\ Slaybaugh is with the Department
of Nuclear Engineering, University of California, Berkeley.}
\thanks{Manuscript received June 31, 2017}}

\title{Gnowee: A Hybrid Metaheuristic Optimization Algorithm for Constrained, Black Box, Combinatorial Mixed-Integer Design} 

\author{
\vspace{20mm}
\\James~E.~Bevins$^{\text{a},\ast}$   R.\ N.\ Slaybaugh$^{\text{b}}$ \\[4pt] 
\textit{$^a$Air Force Institute of Technology, Department of Engineering Physics}\\[-10pt]       
\textit{2950 Hobson Way, Wright-Patterson Air Force Base, OH 45433, USA} \\[-5pt]
\textit{$^b$University of California, Berkeley, Nuclear Engineering Department} \\ [-10pt]
\textit{4173 Etcheverry Hall, Berkeley, CA 94720, USA} \\ [-2pt]
{$^\ast$Email: \href{mailto:james.bevins@afit.edu}{james.bevins@afit.edu}}}       

\date{                               
\vspace{40mm}
Number of pages: \pageref{LastPage} \\
Number of tables: \totaltables \\
Number of figures: \totalfigures \\}

\maketitle

\pagebreak
\begin{abstract}
{This paper introduces Gnowee, a modular, Python-based, open-source hybrid metaheuristic optimization algorithm (Available from \url{https://github.com/SlaybaughLab/Gnowee}).
Gnowee is designed for rapid convergence to nearly globally optimum solutions for complex, constrained nuclear engineering problems with mixed-integer and combinatorial design vectors and high-cost, noisy, discontinuous, black box objective function evaluations.
Gnowee's hybrid metaheuristic framework is a new combination of a set of diverse, robust heuristics that appropriately balance diversification and intensification strategies across a wide range of optimization problems.

This novel algorithm was specifically developed to optimize complex nuclear design problems; the motivating research problem was the design of material stack-ups to modify neutron energy spectra to specific targeted spectra for applications in nuclear medicine, technical nuclear forensics, nuclear physics, etc.  
However, there are a wider range of potential applications for this algorithm both within the nuclear community and beyond.
To demonstrate Gnowee's behavior for a variety of problem types, comparisons between Gnowee and several well-established metaheuristic algorithms are made for a set of eighteen continuous, mixed-integer, and combinatorial benchmarks.
These results demonstrate Gnoweee to have superior flexibility and convergence characteristics over a wide range of design spaces.
We anticipate this wide range of applicability will make this algorithm desirable for many complex engineering applications.

Keywords: Evolutionary computation, metaheuristic, optimization algorithms
}
\end{abstract}

\pagebreak

\section{Introduction} \label{sec:intro}
Many engineering problems and processes can be described by multi-dimensional surfaces that form the ``fitness" landscape of the solution \cite{Lee2011,Yang2010b,Moller2001,Do2011}.  
A single point on the fitness landscape is obtained through evaluation of one n-dimensional solution set against desired objectives such as performance (in the example case, degree of spectrum matching), size, cost, etc. 
All solution sets that meet the given design constraints of a problem form the complete fitness landscape, which can have many locally-optimum solution sets (multi-modal).

One example that motivated this research is the challenge of designing a material stack-up to efficiently modify a neutron energy spectrum from one produced by an available neutron source to one that is desired for a specific application subject to the physical constraints of the implementation of such a stack-up design at the facility.
The ability to develop custom neutron spectra would potentially address capability gaps faced by nuclear forensics, medical isotope production, medical physics, and nuclear data communities, to name a few. 
However, The combination of a large n-dimensional search space, a noisy, multi-modal fitness landscape, and a complex and/or expensive objective function evaluation, often via a physics solver code like Monte Carlo simulations, makes the determination of the globally optimum solution virtually impossible via intuition or parametric studies.
All or many of these characteristics are true for a variety of nuclear design challenges, and formal optimization techniques offer an opportunity to improve the solutions available for these types design challenges while simultaneously increasing the rate of research and development.

At its most basic, an optimization problem can be stated as

\begin{myalign}
  &\substack{\text{Minimize}\\\vv{x} \in \mathbb{R}^d}& f_i&(\vv{x}), & (i&=1,\ 2,\ \ldots,\ I)^T, \label{eq:Obj_Funct} \\
  &\text{Subject to:} & g_j&(\vv{x})\le 0, & (j&=1,\ 2,\ \ldots,\ J), \label{eq:Ineq_Const} \\
  & & h_k&(\vv{x})= 0, & (k&=1,\ 2,\ \ldots,\ K), \label{eq:Eq_Const}
\end{myalign}

\noindent where I, J, and K are the number of objective functions, inequality, and equality constraints, respectively. 
$\vv{x}$ is the candidate design solution given by $\vv{x}=\left(x_1, x_2, \ldots ,x_n \right)^T \in \mathbb{R}^d$, where $\mathbb{R}^d$ is the design space.  
Optimization problems can be classified by sub-setting the mathematical formulation in ways that are illustrative to assessing the requirements for a given optimization algorithm.  
Some of the more general classifications are considered here: single objective ($I=1$) versus multi-objective ($I>1$); linear versus nonlinear objective function(s); differentiable versus derivative free objective function(s); unconstrained ($J=K=0$) versus constrained ($J|K>0$); continuous, discrete, combinatorial, or mixed-integer (MI) design space ($\mathbb{R}^d$); and uni-modal (convex) versus multi-modal (non-convex) fitness landscapes ($f(\vv{x})$) \cite{Guler2010}. 

In this work, we develop an algorithm designed for nuclear engineering applications that fall into the subset of the single objective, nonlinear, derivative-free, constrained or unconstrained, combinatorial  \underline{\textit{and}} mixed-integer, multi-modal categorizations of optimization problems. 
These nuclear engineering applications typically have black box objective functions and/or constraint evaluations, may not be continuous, and can be very noisy due to limitations in the model or underlying nuclear data.
While the categorizations addressed by this work cover most of the problem space for nuclear engineering applications and design problems of interest, there are many applications that could benefit from or require considering multiple objective functions.
This is not yet explicitly considered for the algorithm presented here.

The literature is filled with examples of optimization algorithms designed to solve problems with either continuous, discrete, or combinatorial design vectors \cite{Lee2011, Storn1997, Marichelvam2014, Yang2009, Ouaarab2014, Yang2014, Hou2016}.  
For many engineering problems of interest, this approach can be valid directly for limited cases with one design variable type, or made to work through mapping discrete variables onto continuous space or subsetting combinatorial variations.
However, many practical engineering problems would benefit from a combined approach to fully sample the design space.  
For example, consider the design of a composite shield for mixed radiation \cite{Hu2008}.  
The full design variable can be given by

\begin{equation}
  \vv{x}=\left(x_1, x_2, x_3 \right),
\end{equation}

\noindent where $x_1$ is the $N \times M$ matrix of $N$ materials with up to $M$ elemental components, $x_2$ is the vector of ordering for the $N$ composite materials, and $x_3$ is the vector of thicknesses for $N$ composite materials.  

Hu et al. \cite{Hu2008} made the radiation shielding problem continuous and solved the optimization problem stepwise by first optimizing $x_1$ and then optimizing $x_3$ for a small subset of layered material combinations.  
However, a more robust approach that samples the full solution phase space would treat the optimization of continuous $x_3$, combinatorial $x_2$, and discrete $x_1$ concurrently. 
We assert that this concurrent treatment would increase design automation, provide better solutions, and achieve them more quickly than current approaches. 

The field of mixed-integer nonlinear programming (MINLP) most closely addresses this problem \cite{Burer2012, Floudas2009}.
Many techniques have been employed to solve MINLP through relaxation of the problem space and constraints, convexation of the problem, or application of problem-specific deterministic approaches, but these are not generally applicable or successful when faced with black box, noisy, discontinuous, or combinatorial problems \cite{Floudas2009, Bonami2012, Hemker2008}.
Surrogates and response surface approaches have proven quite successful at addressing those limitations, but these approaches are often problem specific and/or still require smooth, continuous fitness landscapes \cite{Hemker2008, Regis2005}.
On the other hand, metaheuristic approaches are adaptable to a wide range of problems, and techniques have been developed to solve MI, continuous, discrete, and combinatorial problems \cite{Yang2014, Egea2014, Yiqing2007, Tao1998}.

The new hybrid metahueristic optimization algorithm, Gnowee, is a nearly globally convergent algorithm capable of handling noisy, black box, nonlinear, and discontinuous objectives and constraints informed by continuous, integer/binary, discrete, and/or combinatorial design spaces.
Gnowee combines elements from many well established algorithms under the construct of a diverse, coherent, metaheuristic framework that rapidly converges to a minimum fitness threshold, but no global convergence is guaranteed or evaluated \cite{Yang2009, Lones2014, Sorensen2016}.
Emphasis is placed on nearly global convergence (as opposed to true global convergence) as Gnowee is developed for complex engineering design challenges with high-cost objective function evaluations. 
For such problems, the differences between the performance of the true global optimum versus a nearly global optimum is likely to be within the precision of the black box model and underlying data used. 
Further, it is not useful to improve design precision beyond the point practical for construction (e.g.\ consider machining tolerances and cost).     

This  paper  is  organized  as  follows.  
Section \ref{sec:background} provides useful background information.  
Section \ref{sec:algorithm} describes the new Gnowee algorithm.   
The benchmark calculation settings and process are described in Section \ref{sec:set-up}, and the benchmark results are presented in Section\ref{sec:results}. 
Finally, Section \ref{sec:conclusion} summarizes the findings obtained using Gnowee and indicates possible directions for future research.

\section{Background}
\label{sec:background}
In this section, a basic overview of metaheuristic optimization is covered.
Next, classifications of metaheuristic search operators (heuristics) are discussed.
Finally, Lévy flights are described for use in developing Markov chains.

\subsection{Metaheuristics Overview}
The term metaheuristic was coined by Glover in 1986 by combining the Greek prefix meta- (metá, beyond in the sense of high-level) with heuristic (from the Greek heuriskein or euriskein, to search) \cite{Sorensen2016, Glover1986}.  
Sorensen and Glover define metaheuristic as ``a high-level problem-independent algorithmic framework that provides a set of guidelines or strategies to develop heuristic optimization  algorithms" \cite{Sorensen2016}.
Originally intended to describe the framework of search strategies, the term metaheuristic is commonly used as being synonymous with nature-inspired stochastic algorithms, of which there are numerous and varied implementations in the literature \cite{Yang2014, Fister2013}.  
For the purposes of this paper, metaheuristics will be used to refer to the search technique framework developed, and heuristic will be used to refer to the specific search technique used.   

While not holding to any specific metaphor of a nature-inspired optimization process, algorithmic implementations for many heuristics used by Gnowee were derived from those common to nature-inspired optimization algorithms.
Traditional details of the processes mimicked by those algorithms are bypassed, and a description of the search strategies and algorithmic implementation are provided instead.  
By following this approach to algorithm development, i.e.\ developing a coherent, synergistic set of search strategies instead of adhering strictly to an optimization metaphor, a more robust optimization algorithm can be obtained while tailoring the resulting convergence characteristics.  
Finally, the No Free Lunch Theorem states that no algorithm will be universally superior, and this approach of identifying the heuristic search strategies allows for quick identification of the strengths and weaknesses of the metaheuristic framework constructed \cite{Wolpert1997}.
The identification of strengths and weaknesses through characterization of the heuristics employed can be used to assess the applicability of an algorithm to a specific problem.

\subsection{Metaheuristic Search Strategies} \label{sec:meta}
Lones \cite{Lones2014} details a list of ten heuristic search operators that can be used to describe and classify metaheuristic-based algorithms. 
While not all inclusive nor describing the many variations of each approach in the literature, it provides a foundation upon which a metaheuristic framework can be built.   
It is beyond the scope of this paper to reproduce or expand upon Lones's results here, but the basic concepts and applicability of the heuristic search strategies are described for completeness.
Additionally, the set of heuristics are included as a guide to understand the design decisions made through analyzing the inclusion, exclusion, and/or combination of different heuristics within the Gnowee metaheuristic framework.  

The heuristic search strategies presented by Lones are \cite{Lones2014}: 
\begin{itemize}
  \item \textbf{Neighborhood Search:} New solutions developed are a move, or step change, away from current solutions.  A move can be defined as a change in a single variable or the entire solution. While the neighborhood can be defined and sampled multiple ways, the search carried out typically is local in nature. 
  \item \textbf{Hill Climbing:} A sequence of moves is used to find a locally optimum solution where moves are only accepted if they lead to an improvement in fitness.  The acceptance of only positive moves along a search path can lead to local optima trapping if not combined with another global search heuristic.  
  \item \textbf{Accepting Negative Moves:} Allows moves to worse fitness solutions.  Often used to prevent premature convergence at local optima.
  \item \textbf{Multi-Start:} Restart the search process with a new starting location once it converges at a local optimum.  The use of a population can be considered a form of multi-start.  The use of restarts and/or population-based techniques reduces local optima trapping in large search spaces.
  \item \textbf{Adaptive Memory Programming:} Uses past search experience to guide development of the next move.  This can range from calculating random steps from the current design point to stored histories of recent moves.  These strategies can help guide the search away from well explored regions, preventing cyclical patterns.
  \item \textbf{Population-Based Search:} Multiple search processes that are often cooperative and executed in parallel.  In cooperative executions, information sharing can be tightly or loosely coupled.  For large, multi-modal search spaces, this can accelerate global convergence, though it can be inefficient in relatively easy problems.   
  \item \textbf{Intermediate Search:}  Explores the space between two or more high fitness solutions.  This can either lead to exploration of new regions between two local optima or faster local convergence if both solutions reside in the region of the same optimum. Particular implementations of this may not be universal as defining a logical ``middle" between different types of design variables will differ.   
  \item \textbf{Directional Search:} Carries out moves according to identified productive directions of search. Implementation can be guided by gradients, known locations of high fitness solutions, or estimates of the derivatives.  This can lead to faster convergence, but, if not scaled properly for the size and complexity of the problem, can lead to premature convergence.  
  \item \textbf{Variable Neighborhood Search:} Explores different neighborhoods around the current search point.  The definition of different neighborhoods can vary from using moves of variable size to fixed definitions of search neighborhoods.  This can help reduce local trapping and increase search efficiency in large, complex problems.              
  \item \textbf{Search Space Mapping:} Constructs a map or partitions on the search space to guide the search process.  This can aid in the efficiency of convergence but can require knowledge of the problem to devise the correct mapping or partitioning.  
\end{itemize}

It is important to note that there is some overlap in the definitions of these strategies, and a particular algorithmic implementation may be categorized by more than one heuristic approach.
This can be especially true when the algorithms employed were developed without explicit thought to the heuristics employed, which is a common feature of many of the existing metaheuristic algorithmic implementations in the literature.
For clarity, since many specific algorithmic implementations were developed by modifying existing concepts within the literature, all possible heuristic classifications will be listed for each, with the primary listed first, when describing the algorithm's heuristics in Section \ref{sec:algorithm}.

\subsection{Lévy Flights}
Metaheuristic implementations are often stochastic, with a preponderance of implementations sampling from variations of uniform or normally distributed processes.  
As many algorithms are built on metaphors of physical, nature-inspired processes, this is a convenient choice as there are many naturally occurring phenomena that follow a normal distribution.
An alternative is the Lévy distribution, which also has a basis in animal foraging behavior \cite{Levy1994, Zhao2015, Tran2004}.
Lévy Flights are implemented as the basis for the random walk processes in the Cuckoo Search (CS) optimization algorithm and its variants \cite{Yang2009}. 

\begin{figure*}[!t]
  \centering
  \includegraphics[width=3.0in]{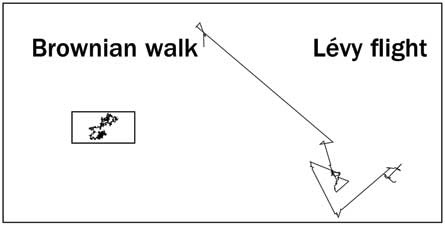}
  \caption{Comparison of equal length paths for a Lévy flight and a normally distributed Brownian walk \cite{Witze2010}.}
  \label{fig:levy_flight}
\end{figure*}

The Lévy distribution is given by 

\begin{equation}
  L_{\alpha,\gamma}(z)=\frac{1}{\pi}\int \limits_{0}^{+\infty} e^{-\gamma q^{\alpha}} \cos(qz) dq,
\end{equation}

\noindent where $\alpha$ defines the index of the distribution and $\gamma$ selects the scale of the process. 
Lévy distributions are heavy-tailed probability distributions where the wings are characterized by a power law behavior.   
This enables a more efficient global search of the phase space to be conducted through a higher probability of medium and large step sizes. 
Not only does the Lévy flight cover more of the phase space, it also avoids revisiting the same design point multiple times.  
This is shown graphically in \autoref{fig:levy_flight} for a Lévy flight and Brownian walk of the same path length.  
       
Several algorithms have been developed to draw a stochastic sample with the correct probability density from a Lévy stable distribution.
In this work, we chose the Mantegna algorithm for its speed and accuracy in sampling a Lévy distribution \cite{Pantaleo2009, Levy1994}.
We show the key steps for the original, full algorithm, as some implementations vary in the literature \cite{Levy1994, Yang2009}.  
First the stochastic variable, $\nu$, is calculated by 

\begin{equation} \label{eq:Levy_1}
  \nu=\frac{x}{y^{\frac{1}{\alpha}}},
\end{equation}

\noindent where $x$ and $y$ are normally distributed stochastic variables with standard deviations $\sigma_x$ and $\sigma_y$.
Since $\sigma_x$ and $\sigma_y$  cannot be chosen independently for arbitrary values of $\alpha$, $\sigma_y=1$ is chosen and 

\begin{equation}
  \sigma_x(\alpha)=\left[ \frac{\Gamma(1+\alpha)\sin(\pi \alpha/2)}{\Gamma \left( ((1+\alpha)/2 \right)\alpha2^{(\alpha-1)/2}} \right]^{\frac{1}{\alpha}}.
\end{equation}  

\noindent A nonlinear transformation can be done to speed up the convergence to a Lévy stable process by

\begin{equation}
  w=\nu \left[ (K(\alpha)-1) e^{(-\nu/C(\alpha))}+1 \right],
\end{equation}

\noindent where $K(\alpha)$ is given by

\begin{equation}
  K(\alpha)= \frac{\alpha \Gamma((\alpha+1)/2\alpha)}{\Gamma(1/\alpha)} \left[ \frac{\alpha \Gamma((\alpha+1)/2)}{\Gamma(1+\alpha)\sin(\pi \alpha/2)} \right]^{\frac{1}{\alpha}},
\end{equation}   

\noindent and $C(\alpha)$ is obtained by numerically solving  

\begin{align}
  \frac{1}{\pi \sigma_x}&\int^{\infty}_0 q^{1/\alpha}\exp \left[-\frac{q^2}{2}-\frac{q^{2/\alpha}C(\alpha)^2}{2\sigma_x^2(\alpha)} \right] dq \nonumber \\ 
  &=\frac{1}{\pi} \int^{\infty}_0 \cos \left[ \left( \frac{K(\alpha)-1}{e}+1 \right)C(\alpha) \right] e^{-q^{\alpha}} dq.
\end{align}

\noindent The stochastic Lévy sample is then given by 

\begin{equation}  \label{eq:Levy_6}
  z=\gamma^{1/\alpha} \frac{1}{n^{1/\alpha}}\sum^n_{k=1}w_k,
\end{equation}

\noindent where $n$ is the number of independent stochastic variables. 
 
\section{Proposed Algorithm} \label{sec:algorithm}
In this section, the key features of the Gnowee algorithm are introduced and described. 
An overview of the Gnowee algorithm's framework is presented, followed by details of each of the initialization and update operators.
Each operator describes the specific implementation in Gnowee, the heuristics used by the operator, and any Gnowee-specific variables that can be modified by the user. 

\subsection{Algorithm Framework}
The general algorithm for Gnowee\footnote{Open source and available from \url{https://github.com/SlaybaughLab/Gnowee}. Unless otherwise state, the URLs in this paper are current as of January 24th, 2018.} is presented in \Cref{alg:Gnowee}.
Operators of the algorithm have been simplified in  \Cref{alg:Gnowee} for conciseness, and each operator is described in detail in the following sections.
Gnowee operators implement the neighborhood search, hill climbing, accepting negative moves, multi-start, adaptive memory programming, population based search, intermediate search, directional search, and variable neighborhood search heuristics.
Search space mapping was not implemented as it generally required  discretization of the phase space.  
A balance was struck between directional search, which can increase the rate of convergence, and accepting negative moves, which can increase the global search capabilities, to accomplish a fast, nearly global convergence within the design criterion described more fully in Section \ref{sec:set-up}.  

\begin{algorithm}[!t] 
  \DontPrintSemicolon
  \SetKwFunction{Initialization}{Initialization}   
  \SetKwFunction{DiscLevyFlight}{DiscLevyFlight}  
  \SetKwFunction{ContLevyFlight}{ContLevyFlight}  
  \SetKwFunction{FitCalc}{FitCalc}  
  \SetKwFunction{Mutation}{Mutation}  
  \SetKwFunction{Inversion}{Inversion}  
  \SetKwFunction{Crossover}{Crossover}
  \SetKwFunction{AOp}{AOp}  
  \SetKwFunction{ThreeOpt}{ThreeOpt}
  \SetKwFunction{Abandon}{Abandon}  
  \SetKwInOut{Input}{Input}\SetKwInOut{Output}{Output}
  \Input{User defined objective function, $f$; constraints, $g$ and $h$; design space, $\vv{x}$; and algorithm settings (see \autoref{table:Gnowee_Setting})}
  \Begin{
    $P.\vv{x}$ $\gets$ initialization($n$)  \tcp*[r]{$P$ is the parent population of size $n$}
    $P.fit \gets$ population$\_$update($P.\vv{x}$) \tcp*[r]{$fit$ is the assessed fitness}
    \While{convergence criterion is not met}{
      $C.\vv{x}$ $\gets$ three$\_$opt($P.\vv{x_x}$)\;
      $P.fit \gets$ population$\_$update($C.\vv{x}$)\;
      \;
      $C.\vv{x_c}$ $\gets$ cont$\_$lévy$\_$flight($P.\vv{x_c}$) \tcp*[r]{$C$ is the child population and $\vv{x_c}$ is the subset of the design vector containing continuous variables}
      $C.\vv{x_d}$ $\gets$ disc$\_$lévy$\_$flight($P.\vv{x_d}$) \tcp*[r]{$\vv{x_d}$ is the subset of the design vector containing discrete, integer, and binary variables}
      $C.\vv{x_x}$ $\gets$ comb$\_$lévy$\_$flight($P.\vv{x_x}$) \tcp*[r]{$\vv{x_x}$ is the subset of the design vector containing combinatorial variables}
      $P.fit \gets$ population$\_$update($C.\vv{x}$, $mh$) \tcp*[r]{$mh$ indicates a Metropolis-Hastings algorithm is to be used}
      \;
      $C.\vv{x}$ $\gets$ crossover($P.\vv{x_c}$, $P.\vv{x_d}$)\;
      $P.fit \gets$ population$\_$update($C.\vv{x}$)\;
      \;
      $C.\vv{x}$ $\gets$ scatter$\_$search($P.\vv{x_c}$, $P.\vv{x_d}$)\;
      $P.fit \gets$ population$\_$update($C.\vv{x}$)\;
      \;
      $C.\vv{x}$ $\gets$ mutation($P.\vv{x_c}$, $P.\vv{x_d}$)\;
      $P.fit \gets$ population$\_$update($C.\vv{x}$)\;
      \;
      $C.\vv{x}$ $\gets$ inversion$\_$crossover($P.\vv{x_c}$, $P.\vv{x_d}$, $P.\vv{x_x}$)\;
      $P.fit \gets$ population$\_$update($C.\vv{x}$)\;
      \;
      $C.\vv{x}$ $\gets$ two$\_$opt($P.\vv{x_x}$)\;
      $P.fit \gets$ population$\_$update($C.\vv{x}$)\;
    }
 }
 \caption{Gnowee Algorithm} \label{alg:Gnowee}
\end{algorithm}

\subsection{Operators}
In this section, each of the major operators in the Gnowee algorithm shown in \Cref{alg:Gnowee} are described.  
For each operator, the basis in the literature, the specific heuristics employed, the Gnowee-specific implementation, and the associated Gnowee user-selected variables are described. 
Recommended values for these variables are shown in \autoref{sec:model_params}.
For the combinatorial operators, a generic, generally applicable approach is described, and areas where slight modifications can be made based on specific optimization applications are noted.  

\subsubsection{Initialization}
The Gnowee algorithm has flexible user selected initialization through random draws from a uniform distribution, nearly orthogonal Latin hypercube (NOLH), NOLH with random permutation, NOLH with Cioppa and Da Rainville permutations, and Latin hypercube (LHC) \cite{Viana2013, Cioppa2007, Rainville2012}.
The use of LHC and NOLH sampling techniques increases the diversity of initial solution sets by better covering the design space.
Increased diversity in the population improves global search while minimizing premature local convergence.
This can dramatically improve early time (i.e.\ few function evaluations) convergence rates, but at late time, only minor differences were noted.
However, improvements in the early time convergence can be useful for the high cost objective function evaluation applications for which Gnowee was designed, and LHC was selected as the default initialization method.

The efficiency of the LHC algorithm to sample the design space is dictated by the number of population members.
Larger populations provide a more robust sampling, but are cumbersome to carry throughout the optimization process due to the increased number of function evaluations performed on ``dead weight" members (see \autoref{sec:model_params} for further discussion on the selection of the population size).
Here, a compromise is struck to allow for robust LHC sampling by generating a number of LHC samples equal to the greater of twice the population size or three times the length of the design vector.
After evaluating the fitness of the candidate designs, only the best $p$ designs are carried forward into the optimization process. 
 
\subsubsection{3-opt}
The 3-opt is in the family of heuristics introduced by Lin and Kernighan that has been applied to a wide variety of different algorithms\cite{Lin1973}.
For the Gnowee algorithm, the 3-opt heuristic is ideal as it is largely problem independent--requiring minimal, if any, knowledge of the problem--and can be implemented rather generically.  
Three-opt implements the neighborhood search, adaptive memory programming, hill climbing, and population-based search heuristics.
To explain 3-opt, consider the following non-looping parent sequence:

\vspace{0.2cm}

\centerline{
$P_1=$ \raisebox{-.3\height}{
\begin{tikzpicture}
\begin{scope}
    \Sequence{A/white!20,H/white!20,B/white!20,D/white!20,G/white!20,F/white!20,C/white!20,E/white!20} .;
\end{scope}
\end{tikzpicture}}}

\vspace{0.2cm}

Three random, unique break points are selected from within the parent.
For this example, $H$, $G$, and $C$.  
To simplify the diagram, consider the following sequence:

\vspace{0.2cm}

\centerline{
$P_1=$ \raisebox{-.3\height}{
\begin{tikzpicture}
\begin{scope}
    \Sequence{S1/white!20,S2/white!20,S3/white!20,S4/white!20} .;
\end{scope}
\end{tikzpicture}}}

\vspace{0.2cm}

\noindent where $S1$ is the sub-sequence given from $A$ to the first break point ($A-H$), $S2$ is the sub-sequence given from the point following the first break point, $B$, through the second break point ($B-D-G$), $S3$ is the sub-sequence given from the point following the second break point, $F$, through the third break point ($F-C$), and $S4$ is the renaming sub-sequence ($E$).
Gnowee considers two re-orderings of the parent sequence around the three chosen break points. 
The first child, $C_1$ is defined as 

\vspace{0.2cm}

\centerline{
$C_1=$ \raisebox{-.3\height}{
\begin{tikzpicture}
\begin{scope}
    \Sequence{S1/white!20,S3/white!20,S2/white!20,S4/white!20} .;
\end{scope}
\end{tikzpicture}}}

\vspace{0.2cm}

\noindent If the fitness of $C_1$ improves upon the fitness of $P_1$, then $C_1$ replaces $P_1$ in the parent population.  
The second permutation is then determined to be

\vspace{0.2cm}

\centerline{
$C_2=$ \raisebox{-.3\height}{
\begin{tikzpicture}
\begin{scope}
    \Sequence{S1/white!20,S2$^{-1}$/white!20,S3$^{-1}$/white!20,S4/white!20} .;
\end{scope}
\end{tikzpicture}}}

\vspace{0.2cm}

\noindent where $S2^{-1}$ and $S3^{-1}$ is the inverse of the original sub-segment.  
If $C_2$ improves the fitness, $C_2$ replaces $P_1$ in the parent population.
This continues for each parent in the initial population. 

\subsubsection{Continuous Lévy Flight}
Lévy flights are used to develop Markov chains that sample the design space.  
Lévy flights have primarily been employed in CS algorithms, but stochastic sampling of the search space is a commonly employed heuristic \cite{Yang2014}.   
Lévy flights implement the variable neighborhood search, adaptive memory programming, hill climbing, directional search, accepting negative moves, multi-start, and population-based search heuristics described in \autoref{sec:meta}.  
Lévy flights on continuous variables are performed on a fraction of parents, $f_l$, as

\begin{equation} \label{eq:Levy_update}
  \vv{x}_r^{g+1}=\vv{x}_r^{g}+ \frac{1}{\beta} \vv{L}_{\alpha,\gamma},
\end{equation}

\noindent where $g$ is the generation number, $r$ is a unique random index, $r \in \{ 1,2,\ldots, p \}$, $\beta$ is a step size scaling factor, and the procedure to determine $\vv{L}_{\alpha,\gamma}$ was described in \autoref{eq:Levy_1} - \autoref{eq:Levy_6}. 
Including $\beta$ is typically necessary to avoid consistently taking steps that are large relative to the dimensions of the problem, which can result in oscillatory behavior.  
The recommended value for $\beta$ reported in the literature varies from $\beta=1$ to $\beta=L/100$ where $L$ is the characteristic scale of the problem being considered \cite{Yang2009, Yang2014, Zhou2014}.
This can be problematic when the scales vary significantly for different variables of the design vector, a fact often ignored.

Instead of scaling $\beta$ to the characteristic scale of the problem, the Gnowee algorithm takes a slightly different approach.
The value of the optimal $\beta$ under this approach was determined using a hyper-optimization technique described in \autoref{sec:model_params}.

For each Lévy flight, the updated design vector is calculated using \autoref{eq:Levy_update}. 
Next, a boundary rejection and re-sampling algorithm is implemented to check the resulting design vector's validity. 
If the step returns a valid result within the boundary of the problem for a given variable of the design vector, the value is accepted and no further action is taken.
However, if that step is outside the problem boundary for a given design variable, that solution is rejected and a new $L_{\alpha,\gamma}$ is generated.  
This method showed drastic improvement in decreasing the sensitivity of the optimization performance with changes in $\beta$, allowing a constant value to be chosen without any impact on the algorithm's performance. 

Finally, within the Lévy flights heuristic, a form of the Metropolis-Hastings rejection sampling algorithm is implemented \cite{Hastings1970}.  
All children that improve upon the fitness of the parent are automatically accepted.
A fraction, $f_mh$, of the children that would normally be discarded due to lower fitness than their parent are instead compared against another random parent in the population.
If the fitness of the child is an improvement over the random parent, the child replaces that parent.
This technique works well when coupled with Lévy flights.
For small steps, accepting a fraction of the solutions that would normally be discarded increases the local convergence rate through the directional search heuristic.
For large steps, this can increase the global exploration through the acceptance of negative moves and multi-start heuristics.  
However, even with Lévy flights, care must be taken not to set $f_mh$ too large or the population diversity will decrease and result in premature convergence.   

\subsubsection{Discrete Lévy Flight} 
Lévy flights on discrete, integer, and binary variables are performed as described in \autoref{eq:Levy_update}, but $L_{\alpha,\gamma}$ is calculated from

\begin{equation} \label{eq:TLF}
  L_{\alpha,\gamma}=ROUND(TLF_{\alpha,\gamma}*D(x)),
\end{equation}

\noindent where $TLF_{\alpha,\gamma}$ is a truncated Lévy flight on the interval $[0,1]$ and $D$ is a variable describing the scale of the discrete variable, $x$, being considered \cite{Mantegna1994}.
In this way, the TLF can be used to map a Lévy distribution onto a discrete variable by eliminating arbitrarily large steps.  
$D$ can be kept general, such as the relative current location indexes, or problem-specific knowledge relevant to the objective function can be leveraged.

For the Gnowee algorithm, one child is generated per $f_l*p$ parents.
The Metropolis-Hasting algorithm is employed to accept a fraction of the children that do not improve upon their parent's fitness.

\subsubsection{Combinatorial Lévy Flight}
Combinatorial Lévy Flight combines \autoref{eq:TLF} from Discrete Inversion Lévy Flight with an inversion operator.
Inversion operators are common in Genetic Algorithm (GA) implementations, and they have been adopted by other algorithms such as Cuckoo Search (CS) \cite{Tao1998, Zhou2014, Starkweather1991}. 
Combinatorial Lévy Flight implements the neighborhood search, hill climbing, adaptive memory programming, directional search, accepting negative moves, multi-start, and population-based search heuristics described in \autoref{sec:meta} by creating new solutions through inverting a portion of the original solution set.
For example, consider the parent sequence below:  

\vspace{0.3cm} 

\centerline{
$P_1=$ \raisebox{-.3\height}{
\begin{tikzpicture}
\begin{scope}
    \Sequence{A/white!20,H/white!20,B/white!20,D/white!20,G/white!20,F/white!20,C/white!20,E/white!20};
\end{scope}
\end{tikzpicture}}}

\vspace{0.2cm}

The inversion points can be selected randomly, based on current location indexes, or by using a problem-specific quantity related to the minimization of the objective function such as distance.  
For example, if H and G are chosen as the cut points, the child would be:  

\vspace{0.2cm}

\centerline{
$C_1=$ \raisebox{-.3\height}{
\begin{tikzpicture}
\begin{scope}
    \Sequence{A/white!20,H/white!20,G/white!20,D/white!20,B/white!20,F/white!20,C/white!20,E/white!20};
\end{scope}
\end{tikzpicture}}}

\vspace{0.2cm}

For the Gnowee algorithm, one child is generated per $f_l*p$ parents.
The Metropolis-Hasting algorithm is employed to accept a fraction of the children that do not improve upon their parent's fitness.

\subsubsection{Crossover}
Crossover is a common feature of GA and Differential Evolution (DE) algorithms implemented in a variety of manners \cite{Storn1997, Tao1998}.  
Variations of the crossover concept have been applied in Particle Swarm Optimization (PSO) and CS algorithms as well \cite{Kennedy1995, Walton2011}. 
For Gnowee, a version of the implementation adopted by Walton is employed for continuous, integer, and discrete variables, where the child solution is calculated as \cite{Walton2011}

\begin{equation}
  \vv{x}_{r}^{g+1}=\vv{x}_{0}^g+\frac{(\vv{x}_{0}^g-\vv{x}_{r}^g)}{\Phi},
\end{equation}

\noindent where $r$ is a unique random index, $r \in \{ 1,2,\ldots, f_e*n \}$, $f_e$ corresponds to the fraction of parents in the elite subset of high fitness designs, and $\Phi$ is the golden ration given by $\Phi=(1+\sqrt{5})/2$.
The golden ratio is used as it has been found to increase convergence over other choices in the range of $[0,2]$ for similar strategies  \cite{Storn1997, Walton2011}. 

As implemented, this method executes the intermediate search, adaptive memory programming, hill climbing, directional search, and population-based search heuristics described in \autoref{sec:meta}. 
The addition of intermediate and directional search through elitism increases overall population fitness and allows for rapid search of promising localities, but can lead to premature convergence due to the loss of population diversity if $f_e$ is not chosen properly.

\subsubsection{Scatter Search}
Gnowee uses an adaptation of scatter search pioneered by Egea et al. \cite{Egea2010}.
The scatter search approach leverages the information of the population to build variable search spaces based on the relative fitness of selected population members.
In Gnowee, the scatter search heuristic is applied to continuous, integer, binary, and discrete variables.
Scatter search implements intermediate search, variable neighborhood search, adaptive memory programming, hill climbing, and population-based search heuristics described in \autoref{sec:meta}. 
The scatter search heuristic updates the design vector according to 

\begin{equation}
  \vv{x}^{g+1} = \vv{c}_1 + (\vv{c}_2-\vv{c}_1) \vv{r},
\end{equation}

\noindent where $\vv{x}^{g+1}$ is the updated design vector and $\vv{r}$ is a vector of uniformly distributed random variables. 
$c_1$ and $c_2$ are given by

\begin{equation} \label{eq:c1}
  \vv{c_1} = \vv{x^g_i} - \vv{d}(1+\alpha \beta),
\end{equation}

\begin{equation} \label{eq:c2}
  \vv{c_2} = \vv{x^g_i} - \vv{d}(1-\alpha \beta),
\end{equation}

\noindent where $\vv{x^g_i}$ is a current member of the elite sub-population.
$\vv{d}$ is given by

\begin{equation}
  \vv{d} = \frac{\vv{x^g_j} - \vv{x^g_i}}{2},
\end{equation}

\noindent where $\vv{x^g_j}$ is a randomly chosen member of the population.
From \autoref{eq:c1} and \autoref{eq:c2}, $\alpha$ and $\beta$ are calculated as

\begin{equation}
  \alpha = 
    \begin{cases}
      1, & \text{if } i < j \\
      -1, & \text{otherwise.}
    \end{cases}
\end{equation}

\begin{equation}
  \beta = \frac{|j-i|-1}{p-2},
\end{equation}

\noindent where $p$ is the size of the population.

This process is continued for $f_e * p$ parents.
If the generated children have a better assessed fitness than their parents, they are accepted into the population and replace their parents.

\subsubsection{Mutation}
Mutation is another foundational search heuristic of GA and DE algorithms that has been modified and adopted by a wide variety of other approaches \cite{Storn1997, Yang2014, Back1993}.  
The Gnowee algorithm implements mutations for continuous, integer, binary, and discrete variables.
The mutation method implements intermediate search, variable neighborhood search, adaptive memory programming, hill climbing, and population-based search heuristics described in \autoref{sec:meta}. 
The mutation method employed is calculated as 

\begin{equation}
  \textit{\textbf{X}}^{g+1}=\textit{\textbf{X}}^g+r \textit{\textbf{D}} (\textit{\textbf{P}}_1-\textit{\textbf{P}}_2),
\end{equation}

\noindent where $\textit{\textbf{X}}$ is the population solution vectors, $\vv{x}$, $r$ is a uniformly distributed random variable, $\textit{\textbf{D}}$ is a matrix of 0s and 1s where each value is determined through a random draw from a uniform distribution with 0 obtained if $<f_m$ and 1 otherwise, and $\textit{\textbf{P}}_1$ and $\textit{\textbf{P}}_2$ are random permutations of the original $\textit{\textbf{X}}$. 

\subsubsection{Inversion Crossover}
Inversion and crossover is used in GA and DE algorithms to copy portions of one parent into another to create a unique child \cite{Storn1997, Tao1998, Back1993}. 
When combined with elitism, crossover can be used to copy traits of high fitness parents into the population.   
In Gnowee, crossover with elitism for continuous, integer, binary,  discrete, and combinatorial variables implements the intermediate search, adaptive memory programming, hill climbing, directional search, and population-based search heuristics described in \autoref{sec:meta}.
Gnowee uses random research to determine the sub-segment to crossover in order to allow for general applicability.  
This can be kept general, such as the relative current location indexes, or problem-specific knowledge relevant to the objective function can be leveraged \cite{Tao1998, Zhou2014}.

Consider the following sequences: $P_1$, chosen randomly from elite sub-population, and $P_2$, a unique, randomly-chosen parent from the entire population: 

\vspace{0.2cm}

\centerline{
$P_1=$ \raisebox{-.3\height}{
\begin{tikzpicture}
\begin{scope}
    \Sequence{A/white!20,H/white!20,G/white!20,D/white!20,B/white!20,F/white!20,C/white!20,E/white!20};
\end{scope}
\end{tikzpicture}}}

\vspace{0.2cm}

\centerline{
$P_2=$ \raisebox{-.3\height}{
\begin{tikzpicture}
\begin{scope}
   \Sequence{E/white!20,A/white!20,G/white!20,C/white!20,H/white!20,B/white!20,D/white!20,F/white!20};
\end{scope}
\end{tikzpicture}}} 

\vspace{0.2cm}

\noindent Point $H$ is chosen randomly as the first inversion point from $P_1$. 
The second inversion point is determined by the location that follows $H$ in $P_2$, which is $B$ in this example.  
After inversion, the child sequence is

\vspace{0.2cm}

\centerline{
$C_1=$ \raisebox{-.3\height}{
\begin{tikzpicture}
\begin{scope}
    \Sequence{A/white!20,H/white!20,B/white!20,D/white!20,G/white!20,F/white!20,C/white!20,E/white!20};
\end{scope}
\end{tikzpicture}}} 

\vspace{0.2cm}

\noindent Only children that improve upon their parent's fitness are accepted.
Next, the previous second inversion point, $B$, is taken as the starting inversion point in $P_2$, and the new second inversion point is determined from the location that follows $B$ in $P_1$, which is $F$.  
The new child sequence generated is

\vspace{0.2cm}

\centerline{
$C_2=$ \raisebox{-.3\height}{
\begin{tikzpicture}
\begin{scope}
    \Sequence{E/white!20,A/white!20,G/white!20,C/white!20,H/white!20,B/white!20,F/white!20,D/white!20};
\end{scope}
\end{tikzpicture}}} 

\vspace{0.2cm}

\noindent If $C_2$ improves upon the fitness of $P_2$, then the child replaces the parent in the population. 
This process is continued over each location in the elite parent for $f_e * p$ elite parents.  

\subsubsection{2-opt}
Two-opt is one of a family of heuristics introduced by Lin and Kernighan that has been applied to a wide variety of different algorithms \cite{Lin1973}.
The Gnowee algorithm adopts an implementation similar to that described by Zhou by adding Lévy flights and elitism to the selection process \cite{Zhou2014}.  
As implemented, 2-opt employs the neighborhood search, adaptive memory programming, hill climbing, directional search, and population-based search heuristics.

To describe 2-opt, consider the following parent sequence

\vspace{0.2cm}

\centerline{
$P_1=$ \raisebox{-.3\height}{
\begin{tikzpicture}
\begin{scope}
    \Sequence{A/white!20,H/white!20,B/white!20,D/white!20,G/white!20,F/white!20,C/white!20,E/white!20} 
\end{scope}
\end{tikzpicture}}}

\vspace{0.2cm}

\noindent The first break point is chosen as $A$.
The second break point, $G$ for this example, is chosen from a TLF mapped onto the length of the sequence.
The re-connection is made where the first break point connects to the second break point and inverts the sequence between the two points.
The point originally following the first break point, $H$, then connects to the point originally following the second break point, $F$ as shown in the sequence

\vspace{0.2cm}

\centerline{
$C_1=$ \raisebox{-.3\height}{
\begin{tikzpicture}
\begin{scope}
    \Sequence{A/white!20,G/white!20,D/white!20,B/white!20,H/white!20,F/white!20,C/white!20,E/white!20} .;
\end{scope}
\end{tikzpicture}}}

\vspace{0.2cm}

If the child improves upon the parent's fitness, then the solution is accepted.  
The starting break point then cycles through the list up to the $n^{th}$ item in the sequence (or $n - 2$ for non-looping sequence) for each parent in the elite subset given by the top $f_e*p$ parents.

\subsubsection{Population Update} 
The descriptions of the individual operators describe the population update procedures used, but it is worth expounding upon this process.
There are a few typical update strategies.
One common approach is to select $p$ new parents from the generated $c$ children plus $p$ existing parents.
In this approach, the $p$ best solutions are taken regardless of their origin, which can result in faster convergence at the expense of reduced diversity with a corresponding increased chance of premature local convergence.
An alternative approach is to select the new $p$ population members from the $c$ children.
This approach increases diversity and limits premature convergence, but may require more function evaluations as a large number of negative steps can be accepted.

The approach adopted by Gnowee attempts to strike a balance by \textit{mostly} only allowing children to replace their parents.
This keeps diversity high and avoids negative moves.
As noted, however, elitism based heuristics, while serving to increase convergence, can also themselves decrease the diversity of the population.
This risk is offset by employing the Metropolis-Hastings algorithm once per generation to allow for negative moves that, due to the nature of Lévy flights, can increase the diversity of the population.

\section{A Study of Characteristics of Gnowee} \label{sec:set-up}
Gnowee was designed for rapid convergence of nearly globally optimum solutions for complex engineering problems with mixed-integer and combinatorial design vectors and high-cost, noisy, discontinuous, black box objective function evaluations.  
However, no coherent set of benchmarks exist for this problem space, nor are there completely comparable codes against which to test.
Instead, continuous, mixed-integer, and combinatorial implementations of several common metaheuristic algorithms, described in \autoref{sec:bench_algos}, were chosen to benchmark against.
The algorithm settings and convergence criteria used are described in \autoref{sec:model_params}.
Finally common benchmark problems from each class of variables were chosen and are described in \autoref{sec:cont_bench}, \autoref{sec:mi_bench}, and \autoref{sec:comb_bench}.

\subsection{Benchmark Comparison Algorithms} \label{sec:bench_algos}
Several well-established, openly available algorithms were chosen to benchmark the performance of Gnowee.  
This section describes the algorithms chosen, the origin of the specific implementation chosen, and any modifications made to the original algorithm.  

It is beyond the scope of this paper to describe each algorithm implemented for benchmarking in detail.
However, with the exception of Discrete Cuckoo Search (DCS), all algorithms were adopted directly from their authors with modifications only to the convergence criteria, described in \autoref{sec:model_params}, and output information.  
References for the source location of each algorithm is described below.
The algorithms chosen for benchmarking represent a diverse set of approaches with both unique and sometimes overlapping metaheuristic characteristics.  
However, we do not claim the implementations chosen represent the best possible implementation of a given algorithm.  
Instead, preference is given to well characterized algorithms available for use in engineering design problems.
 
For comparison against continuous vector algorithms, GA, simulated annealing (SA), PSO, CS, MCS, and MEIGO were selected.  
GA, SA, and PSO were implemented using the Global Optimization Toolbox in Matlab\textsuperscript{\textregistered} R2015b \cite{Mathworks2015}.
A Matlab\textsuperscript{\textregistered} implementation of Yang's CS\footnote{Available from \url{https://www.mathworks.com/matlabcentral/fileexchange/29809-cuckoo-search--cs--algorithm}} and Walton's MCS\footnote{Available from \url{https://www.mathworks.com/matlabcentral/linkexchange/links/2999-modified-cuckoo-search-mcs-open-source-gradient-free-optimiser}} algorithms were obtained  through the Mathworks\textsuperscript{\textregistered} File and Link exchanges \cite{Yang2009, Walton2011}.
The Matlab\textsuperscript{\textregistered} implementation of MEIGO\footnote{\url{http://www.iim.csic.es/~gingproc/meigo.html}},  openly available from the author's website, was used \cite{Egea2014}.

For comparison against mixed-integer vectors, GA and MEIGO were chosen.
GA was implemented using the Global Optimization Toolbox in Matlab\textsuperscript{\textregistered} R2015b \cite{Mathworks2015}.
A Matlab\textsuperscript{\textregistered} implementation of MEIGO was obtained through the MEIGO website \cite{Egea2014}.

Finally, for combinatorial vector algorithms, Discrete GA and DCS were chosen.
The Discrete GA algorithm for TSP was obtained  through the Mathworks\textsuperscript{\textregistered} File exchange\footnote{Available from \url{https://www.mathworks.com/matlabcentral/fileexchange/13680-traveling-salesman-problem-genetic-algorithm}}.
No available implementation of the DCS algorithm was found, so the authors implemented their best interpretation of the algorithm by Zhou \cite{Zhou2014,Ouyang2013}.  
Results from this implementation are consistent with those reported by Zhou \cite{Zhou2014}.  

\subsection{Discussion of Key Model Parameters} \label{sec:model_params}
Each of the considered comparison algorithms has many possible user directed modifications that can be made through adjustment of algorithm parameters.
For this study, all non-convergence related model parameters for the comparison algorithms are set to the defaults recommended by that software package and/or author \cite{Yang2009, Egea2014, Zhou2014, Walton2011, Mathworks2015}.

The convergence criteria were set to ensure fair comparisons were made among the algorithms.
The maximum number of function evaluations, $F^{eval}_{max}$, is capped at 200,000 for all algorithms.
The stall convergence criteria is set at 1E-6 with the maximum number of stall evaluations, $F^{eval}_{stall}$, of 10,000.  
The fitness convergence is set at 1\% of the known optimal fitness.  
Finally, because the solution process is stochastic, each problem was run 100 times to get the representative mean and standard deviation behavior. 
No attempt was made to control for or quantify run time as the intended applications will have high objective function evaluation costs that dwarf the metaheuristic framework costs. 

The Gnowee program-specific settings are shown in \autoref{table:Gnowee_Setting}.
The parameters were determined from the set of benchmarks using iterative parametric studies.
True hyper-optimization techniques were not used as most of the parameters had a large, stable minimum region making this approach sufficient.  

\begin{figure*}[!t]
\centering
{\includegraphics[width=3.5in]{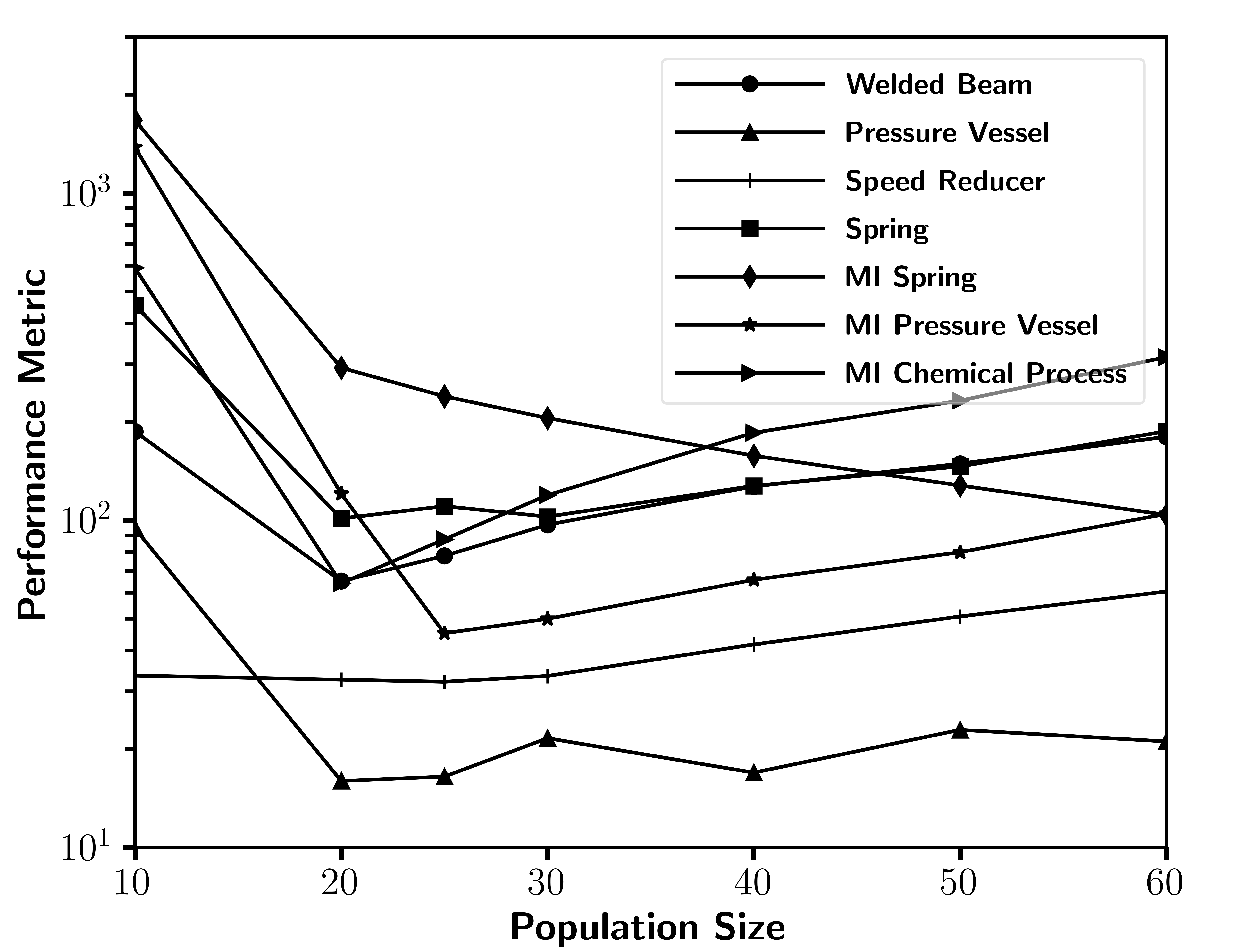}%
\label{p_MI_Constr}
\caption*{Mixed-Integer and Continuous Constrained}}
{\includegraphics[width=3.5in]{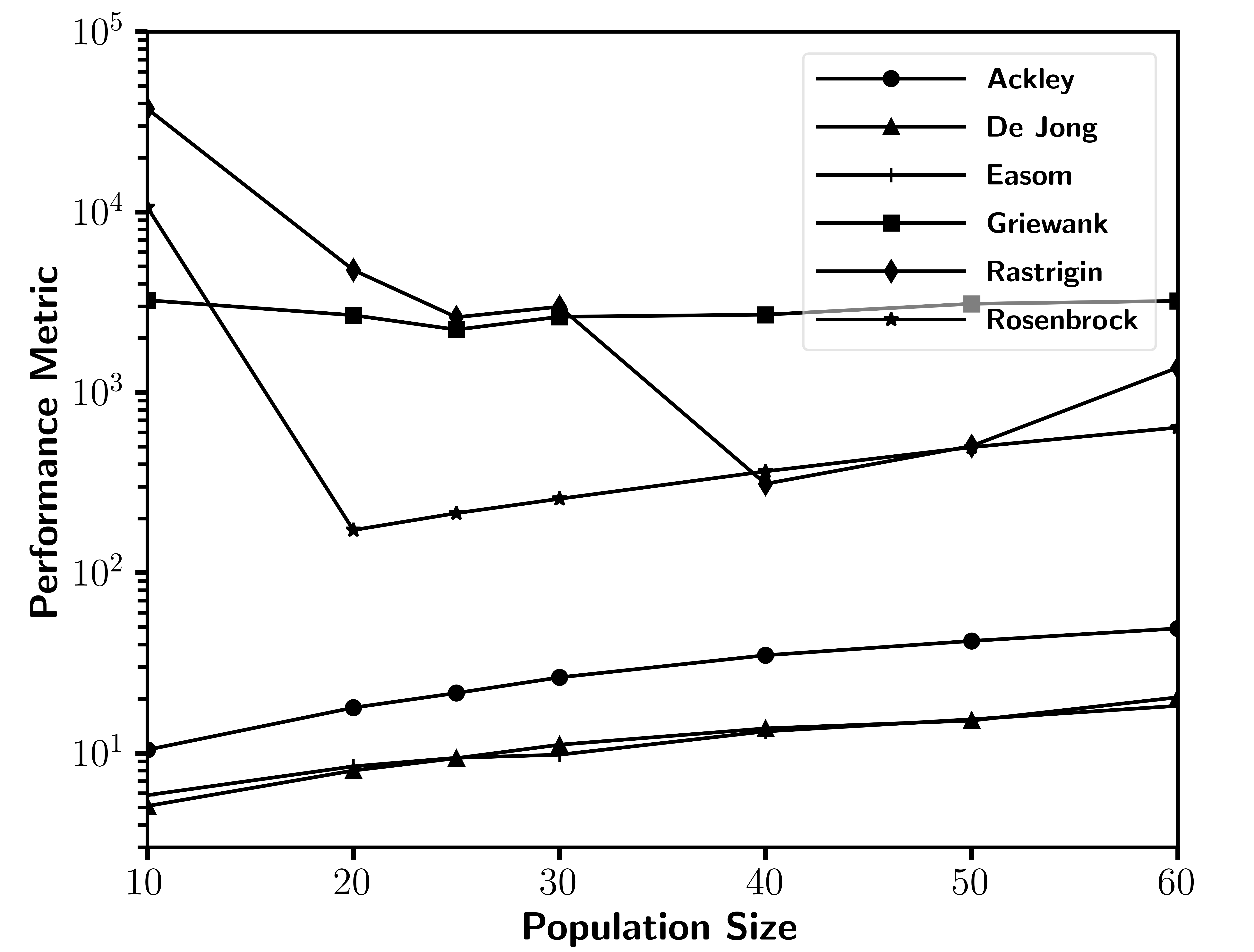}%
\label{p_MI_UnConstr}
\caption*{Continuous Unconstrained}}
{\includegraphics[width=3.5in]{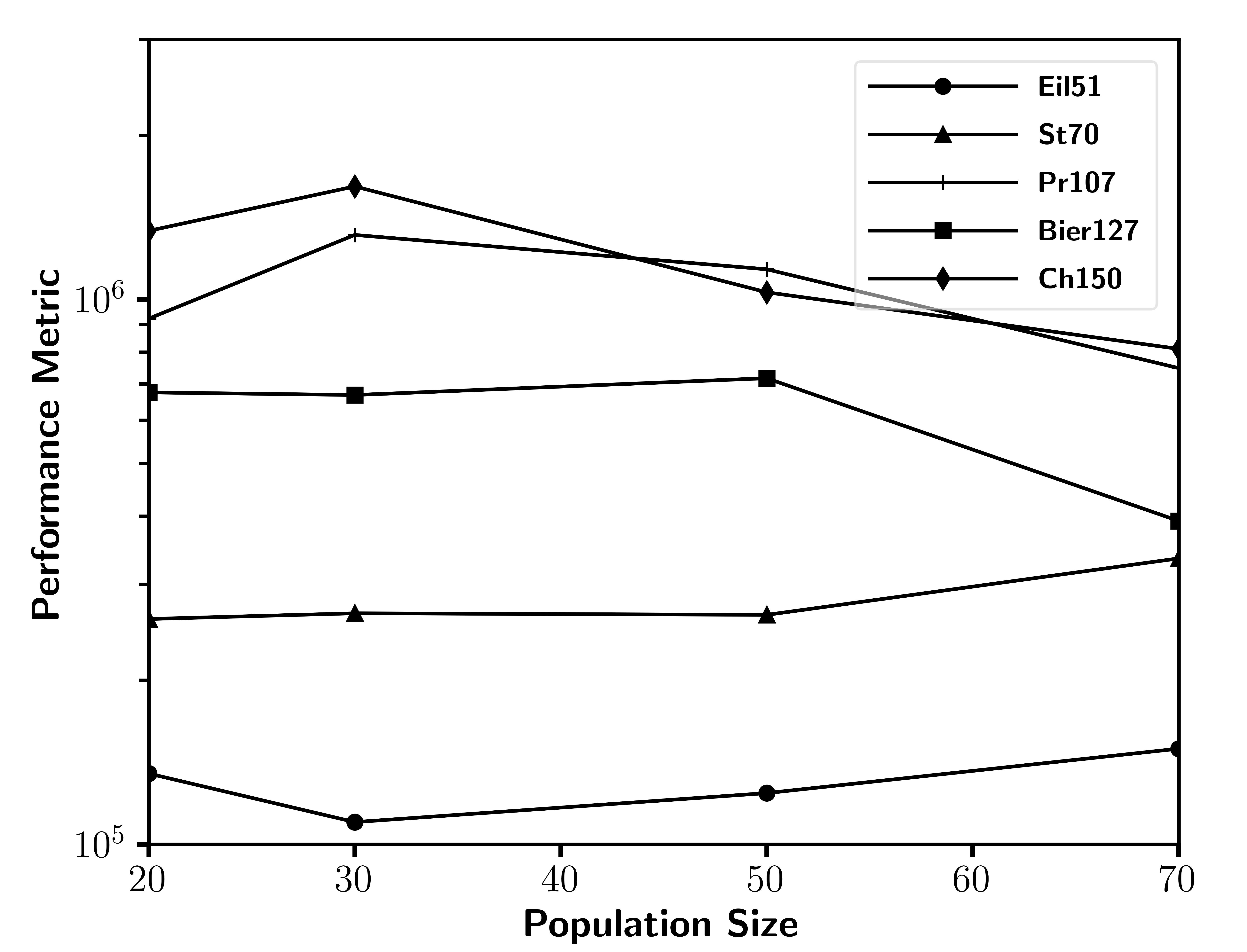}%
\label{p_Comb}
\caption*{TSP}}
\caption{Parametric hyper-optimization results for Gnowee population size. Each algorithm and benchmark were run for 100 iterations.  The performance metric is defined in \autoref{eq:fom}.}
\label{fig:p_hyper}
\end{figure*}

For brevity, only the results from the selection of the population size, $p$, for Gnowee are presented in \autoref{fig:p_hyper} as it is the most sensitive and complex parameter\footnote{Full results for all parameters are available at \url{https://github.com/SlaybaughLab/Gnowee/tree/master/Benchmarks}}. 
As shown in \autoref{fig:p_hyper}, each class of problem, described further in \autoref{sec:cont_bench}, could have a different optimal population size, but 25 was chosen as it maximized benefits for the largest number of considered benchmark problems while minimizing the loss of performance for the remaining problems.  

\begin{table}[!t]
\centering
\caption{Gnowee algorithm settings.} 
\label{table:Gnowee_Setting}
\begin{tabular}{c c }
\toprule
\textbf{Parameter}  & \textbf{Gnowee} \\
\midrule
$\boldsymbol{P}$ & 25 \\ \addlinespace
$\boldsymbol{S_i}$ & LHC \\ \addlinespace
$\boldsymbol{\alpha}$ & 0.5 \\ \addlinespace
$\boldsymbol{\gamma}$ & 1 \\ \addlinespace
$\boldsymbol{\beta}$ & 10 \\ \addlinespace
$\boldsymbol{f_1}$ & 1.0 \\ \addlinespace
$\boldsymbol{f_d}$ & 0.2 \\ \addlinespace
$\boldsymbol{f_e}$ & 0.2 \\ \addlinespace
\bottomrule
\end{tabular}
\end{table}

\subsection{Continuous Benchmark Problems} \label{sec:cont_bench}
Two sets of continuous benchmark problems were chosen. 
The complete description of each problem is not presented, but the mathematical formulation of the objective function, constraints, and the optimal fitness is shown for an example problem from each set.  
References to literature are provided for the remaining benchmarks.   
 
The first set consists of four continuous, constrained engineering problems that are well described in the literature: optimization of a welded beam \cite{Cagnina2008}, pressure vessel \cite{Cagnina2008, Akhtar2002}, speed reducer \cite{Cagnina2008, Akhtar2002}, and spring \cite{Cagnina2008, Yang2014}.
The pressure vessel design problem is stated as

\begin{myalign}
  &\text{Minimize:}& f&(\vv{x})=0.6224 R L t_s + 1.7781 R^2 t_h +3.1611 L t_s^2  \nonumber \\
  & & & \ \ \ \ \ \ \ \ + 19.8621 R t_h^2  \nonumber \\ 
  &\text{Subject to:} & g_1&(\vv{x})=-t_s+0.01932R \le 0 \nonumber \\
  & & g_2&(\vv{x})= -t_h+0.00954R \le 0 \nonumber \\
  & & g_3&(\vv{x})= -\pi R^2 L - \frac{4}{3} \pi R^3 750*1728 \le 0 \nonumber \\
  & & g_4&(\vv{x})= -240+L  \nonumber 
\end{myalign}

\noindent where $\vv{x}$, the upper, and the lower bounds are given by

\begin{myalign}
  \vv{x}&= \left[R, L, t_s, t_h \right] \nonumber \\
  \vv{x}_{lb} &= \left[10, \SI{1E-8}, 0.0625, 0.0625 \right] \nonumber \\
  \vv{x}_{ub} &= \left[50, 200, 6.1875, 6.1875 \right] \nonumber 
\end{myalign}

\noindent and $R$ is the pressure vessel inner radius, $L$ is the length of the pressure vessel, $t_s$ is the thickness of the pressure vessel shell, and $t_h$ is the head thickness.
All variables are treated as continuous.
 
\begin{figure}[!t]
  \centering
  \includegraphics[width=4.5in]{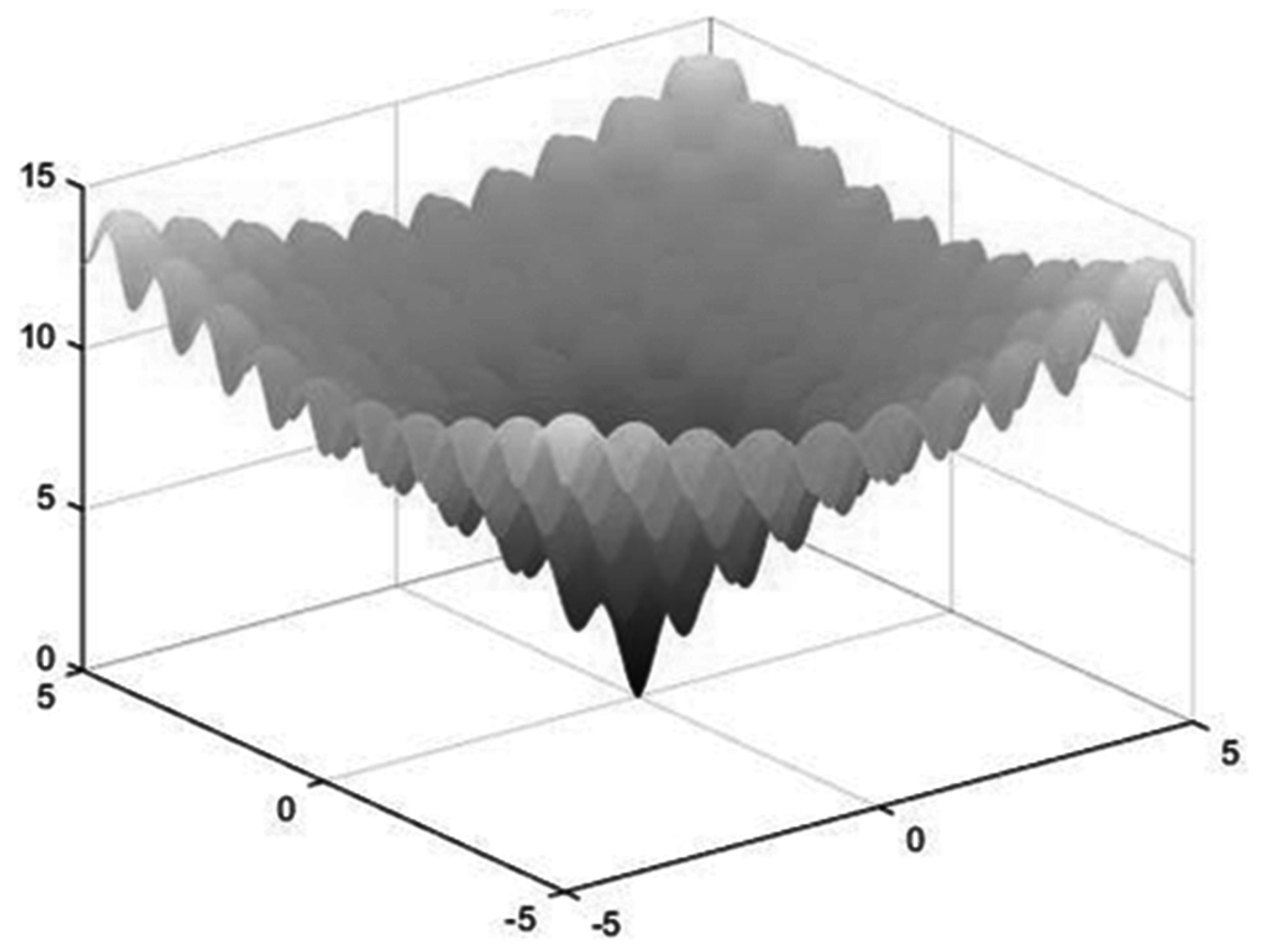}
  \caption{Two-dimensional Ackley function.}
  \label{fig:Ackley}
\end{figure}

The second set consists of six continuous, unconstrained functions that are well described in the literature as benchmarks: Ackley, De Jong, Easom, Griewank, Rastrigin, and Rosenbrock \cite{Yang2014, Mathworks2015, Walton2011}.
Ackley's function is a multi-modal, $n$-dimensional function with a global minimum of $f_{opt}=0$.
It is shown in \autoref{fig:Ackley} for $n=2$ and given by 

\begin{myalign} \label{eq:Ackley}
  f(\vv{x})&=-20exp \left[-0.2 \sqrt{\frac{1}{d} \Sigma_{i=1}^d x_i^2} \right]-exp \left[\frac{1}{d} \Sigma_{i=1}^d cos(2\pi x_i) \right] \nonumber \\
  & + (20+e) \hspace{2.5cm} x_i \in [-32.768,32.768]. \nonumber
\end{myalign}
     
\subsection{Mixed-Integer Benchmark Problems} \label{sec:mi_bench}
Mixed-integer optimization problems combine continuous, binary, integer, and/or discrete variables.
Here, three engineering design benchmarks are considered: MI spring \cite{Lampinen1999}, MI pressure vessel \cite{Cagnina2008}, and a MI chemical process design \cite{Yiqing2007}.
For brevity, only the pressure vessel is described in detail.

The MI pressure vessel design is similar to the continuous pressure vessel design specified in \autoref{sec:cont_bench} with the exception that two variables, $t_s$ and $t_h$, can only take discretized values.
The $\vv{x}$, the upper, and the lower bounds are the same as specified in \autoref{sec:cont_bench}, but for the mixed-integer case, $t_s$ and $t_h$ can only take values that are discrete multiples of 0.0625.

\subsection{Combinatorial Benchmark Problems} \label{sec:comb_bench}
The traveling salesman problem (TSP) was chosen for the combinatorial optimization benchmarks. 
TSP problems were drawn from TSPLIB\footnote{Available from http://comopt.ifi.uni-heidelberg.de/software/TSPLIB95/} and were chosen to span a range of city lengths without being excessively large given the intended engineering design applications.  
For this study, the Eil51, St70, Pr107, Bier127, and Ch150 TSP problems were chosen.

\section{Computational Results and Comparisons} \label{sec:results}
Benchmarks were carried out as described in \autoref{sec:set-up} for continuous constrained, continuous unconstrained, mixed-integer, and combinatorial (TSP) benchmark problems.  
The results presented here highlight the optimization performance of Gnowee against the eight algorithms described in \autoref{sec:bench_algos}.  
For brevity, only a subset of the results are included; the full results set is available from \url{https://github.com/SlaybaughLab/Gnowee/tree/master/Benchmarks/results}. 
Here, specific results are presented for the continuous constrained spring optimization to serve as an example, and summary results are presented for the remaining problems.  

For Gnowee, success is measured by consistent, rapid convergence to nearly globally optimum solutions.  
To benchmark in a method consistent with the Gnowee design objectives, convergence criteria were set to a "limited" number of function evaluations (200,000), a threshold of 1\% of the best known optimal fitness, or a "limited" function evaluation stall limit (10,000).   
To quantitatively capture the performance of a given algorithm to facilitate comparison of algorithm performance, a figure of merit (FOM) is defined as

\begin{equation} \label{eq:fom}
  FOM = \frac{f_{avg}(\vv{x})-f_{opt}}{f_{opt}} \left(N^{avg}_{f(\vv{x})} + 3 \sigma_{N_{avg}} \right)\:.
\end{equation}

This formulation of the FOM emphasizes the convergence rate and width of the the functional evaluation distribution while also penalizing prematurely convergent algorithms for poor fitness solutions.  
The choice of using the $3\sigma$ number of function evaluations emphasizes the importance of minimizing the total number of function evaluations, an important factor for the high-cost objective function evaluation applications for which Gnowee was designed.  
In these applications, an optimization is often only performed a few times, perhaps even just once, making the edge cases more interesting than the average behavior of the algorithm.    

In the results presented below, bolded values indicate premature convergence, where the fitness achieved is $> 1\%$ from the optimal solution.
The italicized values indicate the best performing algorithm according to the FOM.

\subsection{Continuous Constrained Benchmark Results} \label{sec:con_con_results}
The continuous benchmarks considered are subdivided into constrained and unconstrained classes for ease of comparison.  First, the detailed results for the spring optimization problem are shown in \autoref{table:Spring_Detail} for the run conditions and convergence criteria specified in \autoref{sec:model_params}.
\autoref{table:Spring_Detail} presents design variable values ($x_*$), fitness ($f(\vv{x})$), number of function evaluations ($N_{f(\vv{x})}$), and FOM for both the average and best performance obtained over the 100 optimization iterations performed by each algorithm. 
For the spring optimization, Gnowee outperformed the other algorithms on average by reducing the number and standard deviation of function evaluations required, while MEIGO and SA had the best single performance optimizations.    

\pagestyle{empty}
\begin{landscape}
\begin{table*}[!]
  \centering
  \caption{Spring  optimization results (Optimum fitness = 0.012665).}
  \label{table:Spring_Detail}
  	\renewcommand\arraystretch{1.5} 
  	\resizebox{1.0\linewidth}{!}{%
    \begin{tabular}{r c c c c c c c }
    \toprule
    \textbf{} & \textbf{GA\cite{Mathworks2015}} & \textbf{SA\cite{Mathworks2015}} & \textbf{PSO\cite{Mathworks2015}} & \textbf{CS \cite{Yang2014}} & \textbf{MCS\cite{Walton2011}} & \textbf{MEIGO\cite{Egea2014}} & \textbf{Gnowee}  \\
    \midrule
    $x_1^{avg}$ & \textbf{0.060063 $\mypm$ 0.00558}
            & 0.050810 $\mypm$ 0.00090
            & \textbf{0.052298 $\mypm$ 0.00314}  
            & 0.051703 $\mypm$ 0.00127 
            & \textbf{0.058462 $\mypm$ 0.00592} 
            & 0.052817 $\mypm$ 0.00133           
            & \textit{0.051620 $\mypm$ 0.00119} \\ \addlinespace
    $x_2^{avg}$ & \textbf{0.610416 $\mypm$ 0.18392}
            & 0.335680 $\mypm$ 0.02133
            & \textbf{0.375903 $\mypm$ 0.08398} 
            & 0.357026 $\mypm$ 0.03064 
            & \textbf{0.560899 $\mypm$ 0.18946}
            & 0.385194 $\mypm$ 0.03386
            & \textit{0.355059 $\mypm$ 0.02874} \\ \addlinespace
    $x_3^{avg}$ & \textbf{5.461663 $\mypm$ 3.19803}
            & 12.840958 $\mypm$ 1.32526
            & \textbf{11.363353 $\mypm$ 3.39514}
            & 11.560822 $\mypm$ 1.83294
            & \textbf{6.573904 $\mypm$ 3.75091}
            & 9.997362 $\mypm$ 1.53760
            & \textit{11.648626 $\mypm$ 1.72111} \\ \addlinespace
    $f_{avg}(\vv{x})$ & \textbf{0.014398 $\mypm$ 0.00161}
                  & 0.012778 $\mypm$ 0.00002
                  & \textbf{0.012890 $\mypm$ 0.00030}
                  & 0.012771 $\mypm$ 0.00004
                  & \textbf{0.014068 $\mypm$ 0.00156}
                  & 0.012785 $\mypm$ 0.00006
                  & \textit{0.012763 $\mypm$ 0.00002} \\ \addlinespace
    $N_{f(\vv{x})}^{avg}$ & \textbf{16947 $\mypm$ 6379}
                & 5820 $\mypm$ 5085
                & \textbf{7959 $\mypm$ 10523}
                & 16948 $\mypm$ 7174  
                & \textbf{19013 $\mypm$ 13684}
                & 10279 $\mypm$ 9835
                & \textit{4738 $\mypm$ 1836} \\ \addlinespace
     $FOM_{avg}$ & \textbf{4938} & 188.3 & \textbf{702.9} & 323.2 & \textbf{6654} & 376.7 & \textit{79.0} \\ \addlinespace
    \midrule
     $x_1^{best}$ & 0.051341 & 0.050318 & 0.052021 & 0.051230 & 0.051768 & \textit{0.052121} & 0.051092 \\ \addlinespace
     $x_2^{best}$ & 0.348240 & 0.324492 & 0.364483 & 0.345779 & 0.358488 & \textit{0.367169} & 0.342205 \\ \addlinespace
     $x_3^{best}$ & 11.878832 & 13.476806 & 10.886333 & 11.987921 & 11.224128 & \textit{10.703094} & 12.210091 \\ \addlinespace
     $f_{best}(\vv{x})$ & 0.012740 & 0.012715 & 0.012711 & 0.012694 & 0.012705 & \textit{0.012691} & 0.012694 \\    \addlinespace
     $N_{f(\vv{x})}^{best}$ & 7900 & 835 & 27900 & 5225 & 4328 & \textit{1090} & 3813 \\ \addlinespace
     $FOM_{best}$ & 47 & 3.3 & 100.3 & 12.1 & 13.5 & \textit{2.2} & 8.7 \\ \addlinespace
     \bottomrule
     \end{tabular}}
\end{table*}
\end{landscape}
\pagestyle{plain}

The summary of FOM results obtained for the continuous, constrained benchmark problems are shown in \autoref{table:Cont_Con_FOM_Summary}.
In \autoref{table:Spring_Detail} and \autoref{table:Cont_Con_FOM_Summary}, bold results indicate fitness greater than 1\% from the global optimum and italicized results indicate the best performance for the average and the overall best run.
Across all of the benchmarks in this category, Gnowee had the best performance twice and never finished worse than third.
Additionally, Gnowee did not suffer from premature convergence issues that affected all of the other algorithms except CS.
Rapid yet continual convergence across a wide range of problems is a key feature of Gnowee enabled by the diverse set of hybrid heuristics.

\begin{table}[!]
\centering
\caption{Summary of FOM results for continuous constrained optimization benchmarks.} 
\label{table:Cont_Con_FOM_Summary}
\begin{tabular}{r c c c c c }
\toprule
\textbf{} & \textbf{Welded} & \textbf{Pressure} & \textbf{Speed} & \textbf{Spring} \\
& \textbf{Beam \cite{Cagnina2008}} & \textbf{Vessel \cite{Cagnina2008}} & \textbf{Reducer \cite{Cagnina2008}} & \textbf{ \cite{Cagnina2008}}  \\
\midrule
GA\cite{Mathworks2015} & \textbf{106667.6} & \textbf{762.0} & 56.5 & \textbf{4938} \\ \addlinespace
SA\cite{Mathworks2015} & \textbf{5125.8} & \textbf{8488.6} & \textbf{1437.1} & 188.3 \\ \addlinespace
PSO\cite{Mathworks2015} & 329.4 & \textbf{1209.5} & 62.2 & \textbf{702.9}\\ \addlinespace
CS\cite{Yang2014} & 442.6 & 529.4 & 51.3 & 323.2 \\ \addlinespace
MCS\cite{Walton2011} & \textbf{2933.8} & \textbf{1472.6} & 26.0 & \textbf{6654}\\ \addlinespace
MEIGO\cite{Egea2014} & \textit{35.0} & \textbf{1567.4} & \textit{11.7} & 355.3 \\ \addlinespace
Gnowee & 80.6 & \textit{108.4} & 29.7 & \textit{79.0} \\ \addlinespace
\bottomrule
\end{tabular}
\end{table}

\begin{figure}[!t]
  \centering
  \includegraphics[width=4.5in]{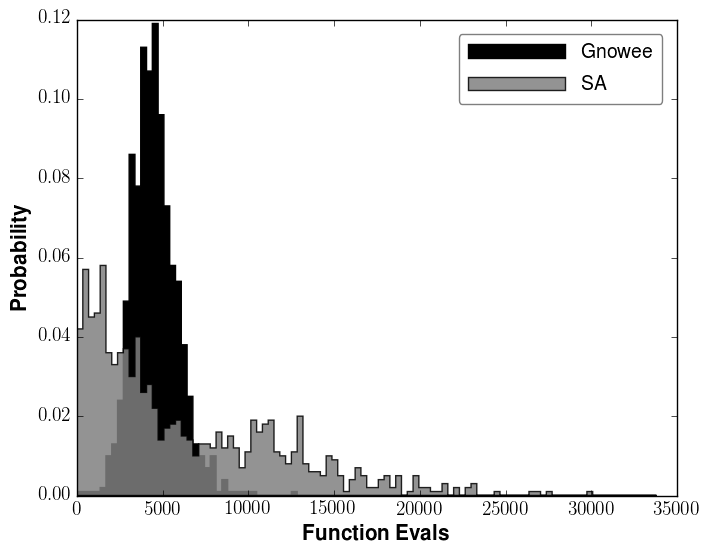}
  \caption{Histogram of the number of function evaluations performed before meeting the convergence or stall criteria specified in \autoref{sec:model_params}.}
  \label{fig:Spring_hist}
\end{figure}

One way to better visualize the results from \autoref{table:Cont_Con_FOM_Summary} is to look at the histogram of converged solutions as shown in \autoref{fig:Spring_hist} for SA and Gnowee.
The histogram includes optimization results for 1000 iterations of the best average performer, Gnowee, and one of the top individual performers, SA.
SA heavily relies on the directional search, neighborhood search, and hill climbing heuristics, resulting in rapid convergence to a \textit{local} optimum.
However, this comes at the expense of robust global search, often resulting in a solution that stalls short of the global, or in this case nearly global, optimum.  
Gnowee, on the other hand, does not achieve nearly as rapid initial convergence, but its convergence is more consistent due to the balancing of local and global search heuristics.  

\begin{figure}[!t]
  \centering
  \includegraphics[width=4.5in]{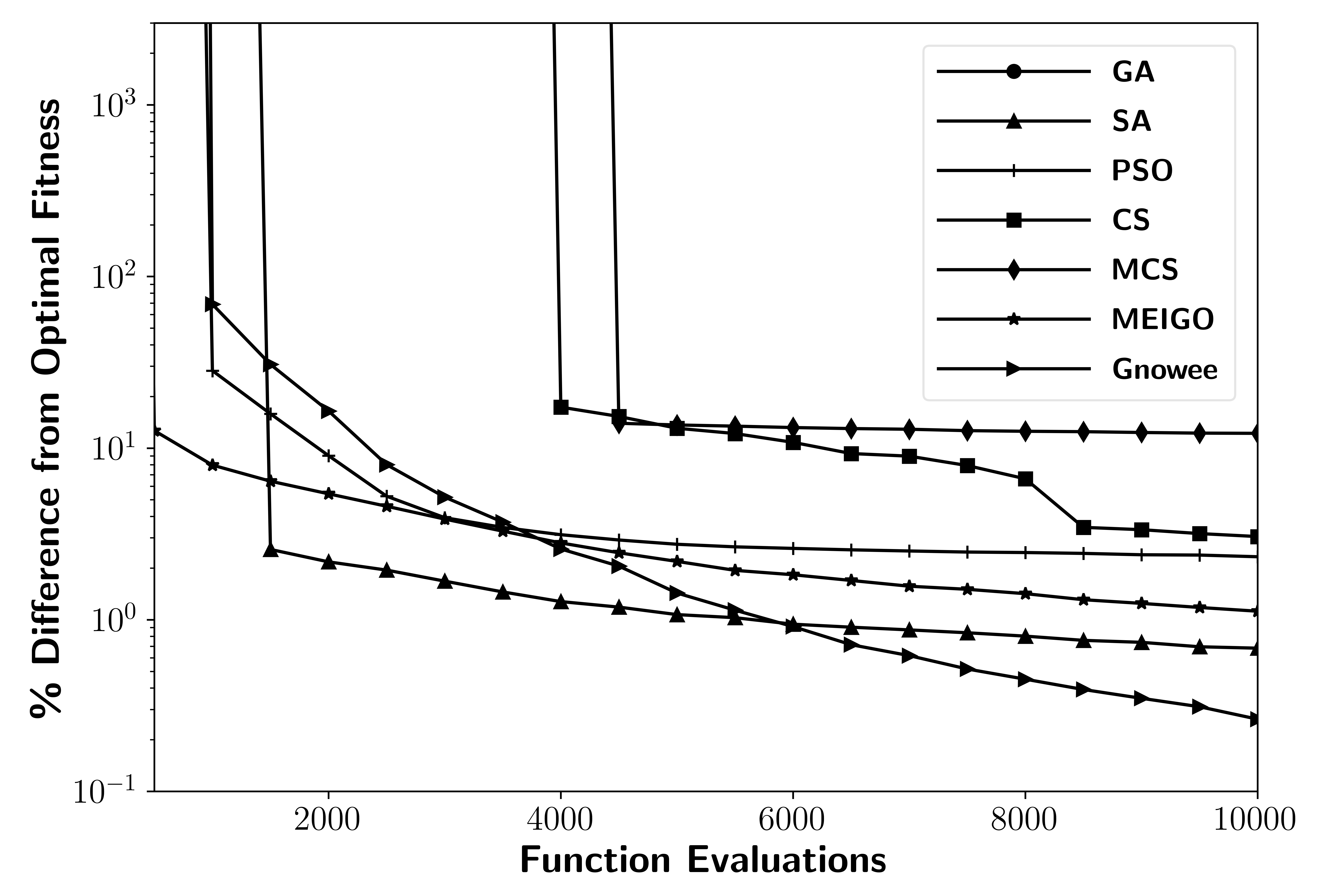}
  \caption{Spring optimization fitness convergence. $1\sigma$ standard deviation in fitness is generally of comparable size or smaller than the plot markers.}
  \label{fig:Spring}
\end{figure}

The effect on the balanced heuristics on the convergence can be shown throughout the complete optimization evolution as illustrated in  \autoref{fig:Spring} for the spring benchmark.
This figure shows the average fitness versus fixed function evaluation intervals for up to 10,000 function evaluations, illustrating some key points.
First, while Gnowee is not the best algorithm at all fitness convergence levels, it is competitive across the full range considered.
Second, Gnowee maintains its convergence rate and continues to converge while other algorithms start to plateau.
Finally, this highlights a potential area for future work.
In this example, and across several others, Gnowee's initial solution is among the worst.
Work to improve the initial starting solution could accelerate Gnowee's convergence and improve performance.

\begin{figure}[!t]
  \centering
  \includegraphics[width=4.5in]{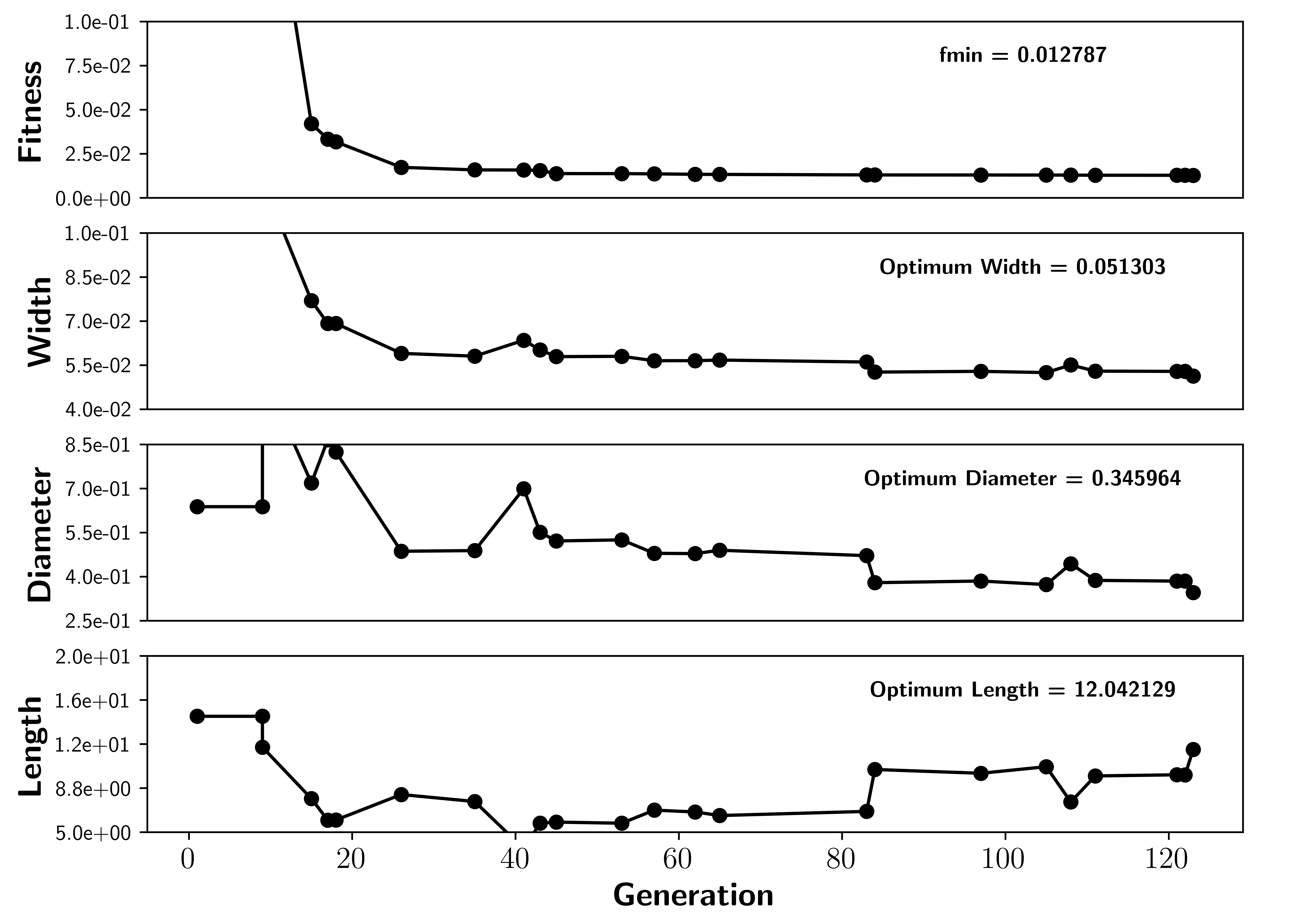}
  \caption{Spring optimization history for a single Gnowee run showing the convergence of the fitness and each design parameter.}
  \label{fig:Spring_param}
\end{figure}

Finally, a look at the convergence history of the design parameters illustrates the importance of balancing local and global search. 
Convergence history is shown in \autoref{fig:Spring_param} for a single optimization of the spring benchmark using Gnowee.  
In \autoref{fig:Spring_param}, each point represents an improvement in overall fitness (top subplot) and the corresponding values for each of the three design variables (bottom three subplots).  
When there are periods of gradual change in the design variables with improvements in fitness, the solution is improving via local search heuristics.
When there are large changes in one or more design variables from one point to the next, the solution is improving via global search heuristics.  

This combination of local and global improvement is the reason that Gnowee outperforms most other algorithms most of the time. 
Unlike GA, SA, PSO, or MCS, which tend to prematurely converge for some problems due to shifts towards local search with increasing generation/evaluations, Gnowee maintains global search heuristics throughout the optimization as illustrated by the jumping between local optima throughout the optimization history shown in \autoref{fig:Spring_param}. 

\subsection{Continuous Unconstrained Benchmark Results} \label{con_uncon_results} 
The other subset of continuous benchmarks was the unconstrained functions described in \autoref{sec:cont_bench}.
The summary of the FOM results obtained for these benchmark problems are shown in \autoref{table:Cont_Uncon_FOM_Summary}, where bold results indicate fitness greater than 1\% from the global optimum and italicized results indicate the best performance for the average and the overall best run.
Each column of the table indicates the benchmark function and the dimensionality considered.  
Here, Gnowee obtains the best performance for the Rastrigin and Rosenbrock functions and the second best performance for the remaining functions.
Similar to the continuous constrained benchmark results shown in \autoref{sec:con_con_results}, Gnowee maintains consistent continual convergence throughout the optimization process for all problems except for the Rastrigin function.
For the easier problems, Gnowee's poor early time solutions slow the convergence process slightly, thereby hindering its performance when compared to the other algorithms. 

\pagestyle{empty}
\begin{landscape}
\begin{table*}[!t]
\centering
\caption{Summary of FOM results for continuous unconstrained optimization benchmarks.}
  	\renewcommand\arraystretch{1.5} 
  	\resizebox{1.0\linewidth}{!}{%
\begin{tabular}{r c c c c c c }
\toprule
\textbf{} & \textbf{Ackley (3-D)} & \textbf{De Jong (4-D)} & \textbf{Easom (2-D)} & \textbf{Griewank (6-D)}  & \textbf{Rastrigin (5-D)} & \textbf{Rosenbrock (5-D)} \\
\midrule
GA\cite{Mathworks2015} & \textbf{1666.8} & 10.7 & \textbf{6809.3} & \textit{92.1} & \textbf{15843} & \textbf{181239} \\ \addlinespace
SA\cite{Mathworks2015} & 109.7 & 53.7 & \textbf{22415.2} & \textbf{27654.5} & \textbf{122362} & \textbf{12420} \\ \addlinespace
PSO\cite{Mathworks2015} & 33.7 & 15.2 & 11.9 & \textbf{3665.4} & \textbf{29436} & \textbf{25884} \\ \addlinespace
CS\cite{Yang2014} & 85.9 & 30.5 & 58.8 & \textbf{13956.8} & \textbf{274627} & \textbf{47472} \\ \addlinespace
MCS\cite{Walton2011} & \textbf{1955.5} & 12.6 & 54.4 & \textbf{16746.4} & \textbf{155804} & \textbf{143587} \\ \addlinespace
MEIGO\cite{Egea2014} & \textit{12.5} & \textit{3.5} & \textit{3.2} & \textbf{2597.3} & \textbf{11514.8} & 224.4 \\ \addlinespace
Gnowee & 19.7 & 8.9 & 6.5 & \textbf{1549.7} & \textbf{\textit{1753.0}} & \textit{152.2} \\ \addlinespace
\bottomrule
\end{tabular}}
\label{table:Cont_Uncon_FOM_Summary}
\end{table*}
\end{landscape}
\pagestyle{plain}

The results in \autoref{table:Cont_Uncon_FOM_Summary} are useful for understanding the interplay of the well defined fitness landscape and the search heuristics used.
For example, consider the Griewank function results.
The Greiwank function is a high-frequency, multi-modal landscape superimposed on a convex design landscape.
When solving this problem, an initial population is likely to be formed in the large convex region and funneled to the edges of the flatter, multi-modal region in the center.

GA's focus on intermediate search heuristics is highly effective in a problem like Griewank.
Intermediate search allows for pairs of solutions at opposite edges to effectively traverse the complex multi-modal region in the center.
Conversely, algorithms that use neighborhood or variable neighborhood search heuristics will have to traverse many local optima that can trap individual members.
Algorithms incorporating intermediate or directional search heuristics such as PSO, MEIGO, and Gnowee do better on this problem.
Algorithms that focus on neighborhood or variable neighborhood search like SA and CS do poorly.
Finally, SA and MCS use "cooling" parameters that limit the range of the neighborhood and variable neighborhood search heuristics as the number of generations progresses.
This can lead to local trapping and premature convergence on large highly multi-modal problems like Griewank and is another factor in their poor performance.   
  
\subsection{Mixed-Integer Benchmark Results}
The mixed-integer benchmarks allow for the performance of Gnowee to be assessed in problems containing a combination of continuous, integer, binary, and discrete variables.
The summary of the FOM results obtained for the mixed-integer benchmark problems are shown in \autoref{table:MI_FOM_Summary}, where bold results indicate fitness greater than 1\% from the global optimum and italicized results indicate the best performance for the average and the overall best run.

\begin{table}[!t]
\centering
\caption{Summary of FOM results for constrained mixed-integer optimization benchmarks.} 
\label{table:MI_FOM_Summary}
\begin{tabular}{r c c c c} 
\toprule
\textbf{} & \textbf{Pressure} & \textbf{Spring \cite{Lampinen1999}} & \textbf{Chemical} \\
& \textbf{Vessel \cite{Cagnina2008}} &  & \textbf{Process \cite{Yiqing2007}}  \\
\midrule
GA\cite{Mathworks2015} & \textbf{897.9} & \textbf{1697.9} & \textbf{18890.1} \\ \addlinespace
MEIGO\cite{Egea2014} & \textbf{358.3} & 330.1  & 95.2 \\ \addlinespace
Gnowee & \textit{40.4} & \textit{219.2} & \textit{92.1} \\ \addlinespace
\bottomrule
\end{tabular}
\end{table}

Here, Gnowee obtains the best performance across all benchmarks considered.
Similar to the previous benchmarks, Gnowee maintains rapid, consistent, continual convergence throughout the optimization process across the range of the benchmarks considered.
Although MEIGO performs similarly well for the spring and chemical process problems, it pre-maturely converges more often than not for the pressure vessel design.
Once again, Gnowee's diverse, robust set of heuristics limits premature convergence across all benchmarks considered.

\subsection{Combinatorial Benchmark Results}

The combinatorial aspects of Gnowee were benchmarked using the TSP problems described in \autoref{sec:comb_bench}.
Unlike the algorithms used to solve the previous benchmarks, the GA and DCS algorithms used TSP specific information, i.e.\ the distance between pairs of cities, to guide the search process.
To enable a coherent comparison between the algorithms, distance based searches were also incorporated into the Gnowee heuristics.

The summary of the FOM results obtained for the TSP benchmark problems are shown in \autoref{table:Comb_FOM_Summary}, where bold results indicate fitness greater than 1\% from the global optimum and italicized results indicate the best performance for the average and the overall best run.
Each row of the table indicates the benchmark TSP route considered.  

\begin{table}[!t]
\centering
\caption{Summary of FOM results for TSP optimization benchmarks.} 
\label{table:Comb_FOM_Summary}
\begin{tabular}{r c c c} 
\toprule
\textbf{}  & \textbf{GA\cite{Mathworks2015}}  & \textbf{DCS\cite{Zhou2014}} & \textbf{Gnowee} \\
\midrule
Eil51 & \textbf{10940.4} & 1509.7 & \textit{\textbf{555.6}} \\ \addlinespace
St70 & \textbf{27452.0} & 3804.6 & \textit{\textbf{1403.1}} \\ \addlinespace
Pr107 & \textbf{36563.1} & \textbf{6067.9} & \textit{\textbf{3380.5}} \\ \addlinespace
Bier127 & \textbf{49869.9} & \textbf{8337.3} & \textit{\textbf{3918.6}} \\ \addlinespace
Ch150 & \textbf{115495.0} & \textbf{14396.6} & \textit{\textbf{5261.4}} \\ \addlinespace
\bottomrule
\end{tabular}
\end{table}

Due to the dimensionality of the TSP problems, indicated by the number in the benchmark problem name, the solution space is extremely large. 
This large space makes 1\% convergence results with limited function evaluations difficult, as illustrated in \autoref{table:Comb_FOM_Summary} by the fact that 13 of the 15 cases do not meet the 1\% convergence criteria.
While not meeting the stated convergence criteria, Gnowee does outperform DCS and GA across all of the TSP benchmarks.

The primary driver for Gnowee's increased performance is a reduction in the number of function evaluations needed, as shown in \autoref{table:Comb_Feval_Summary}.
Gnowee reduces the number of function evaluations by a factor of $\sim$4-10 over the DCS and GA algorithms.
The reduction in function evaluations will enable nearly-global solutions with limited function evaluations in the smaller sets of combinatorial variables found in typical engineering design problems.

\begin{table} [!t]
\centering
\caption{Summary of function evaluation results for TSP optimization benchmarks.} 
\label{table:Comb_Feval_Summary}
\begin{tabular}{r c c c }
\toprule
\textbf{}  & \textbf{GA\cite{Mathworks2015}}  & \textbf{DCS\cite{Zhou2014}} & \textbf{Gnowee} \\
\midrule
Eil51 & \textbf{103415 $\mypm$ 24218} & 104341 $\mypm$ 39398 & \textit{\textbf{9294 $\mypm$ 6033}} \\ \addlinespace
St70 & \textbf{157455 $\mypm$ 30426} & 174362 $\mypm$ 37953 & \textit{\textbf{16238 $\mypm$ 9823}} \\ \addlinespace
Pr107 & \textbf{159692 $\mypm$ 37900} & \textbf{198006 $\mypm$ 15497} & \textit{\textbf{27447 $\mypm$ 16157}} \\ \addlinespace
Bier127 & \textbf{197340 $\mypm$ 10306} & \textbf{228991 $\mypm$ 6345} & \textit{\textbf{37483 $\mypm$ 18077}} \\ \addlinespace
Ch150 & \textbf{199603 $\mypm$ 3276} & \textbf{231786 $\mypm$ 5338} & \textit{\textbf{48757 $\mypm$ 24725}} \\ \addlinespace
\bottomrule
\end{tabular}
\end{table}

\section{Conclusion} \label{sec:conclusion}

This paper presented the new Gnowee algorithm and detailed comparison of its performance to well-established metaheuristic algorithms from the literature.
Gnowee was developed to provide a modular, open-source, general-purpose hybrid metaheuristic optimization algorithm that can achieve rapid, nearly-globally optimum solutions for complex nuclear engineering design challenges.
Gnowee is capable of optimizing constrained, noisy, derivative-free, multi-modal analytic or black box objective functions informed by continuous, integer/binary, discrete, and/or combinatorial variables.
Gnowee is unique in its formulation and broad application where the meta-heurisitic framework enables enables optimization of a wide variety of engineering problems with widely varying design vectors and fitness landscapes.
Additionally, the rapid, consistent convergence of Gnowee enables the algorithm to be employed with high-cost objective functions such as radiation transport calculations.

Gnowee achieves this rapid, nearly global convergence across a large range of design vectors and fitness landscapes through the implementation of a diverse, robust metaheuristic framework that balances diversification and intensification strategies.
This was demonstrated with a set of eighteen benchmark problems spanning multiple variable types where Gnowee's convergence rate and consistency was measured against established metaheuristic algorithms.
While Gnowee was not always the top performing algorithm, it also only finished worse than second once (the speed reducer benchmark where it finished a close third).
Such high performance across so many problem types is why this algorithm is a good choice for many types of nuclear engineering problems. 

Future work might include implementing better strategies to develop the initial population, the addition of population and diversity control algorithms, and a ``bulletin board" approach to parallelization. 
Currently, several initialization strategies are available in Gnowee, but the benchmark results showed Gnowee to consistently generate worse starting solutions than the other algorithms considered.
The results obtained indicate that the metaheuristic framework employed generally maintains a diverse set of solutions, but the few cases that did not converge within the specified criteria often suffered from a loss of diversity.
Actively managing the diversity could improve the algorithm's performance across a wider range of problems.
Finally, Gnowee was implemented with the idea that parallelization would occur at the function evaluation stage following each heuristic.
However, asynchronous, ``bulletin board" approaches could be employed to accelerate the wall time of the solution in time sensitive applications.

Overall, the benchmark results show that Gnowee maintained consistent, continual convergence across a wide set of benchmarks while all of the other algorithms considered suffered from premature convergence issues on several of the benchmarks considered.
Gnowee therefore shows strong promise as an effective and all purpose optimization algorithm for a wide range of nuclear engineering problems.


%



\section*{Acknowledgment}
This material is based upon work supported by the National Science Foundation Graduate Research Fellowship under Grant No. NSF 11-582.

This material is based on work supported by the Department of Energy National Nuclear Security Administration through the Nuclear Science and Security Consortium under Award Numbers DE-NA0000979 and DE-NA0003180. 

This research used the Savio computational cluster resource provided by the Berkeley Research Computing program at the University of California, Berkeley (supported by the UC Berkeley Chancellor, Vice Chancellor for Research, and Chief Information Officer).

\section*{Disclaimer}
The views expressed in this article are those of the authors and do not reflect the official policy or position of the United States Air Force, Department of Defense, or the U.S. Government.


\bibliographystyle{nse}
\bibliography{library}

\begin{thebibliography}{10}
\newcommand{\enquote}[1]{``#1''}
\providecommand{\url}[1]{\texttt{#1}}
\providecommand{\urlprefix}{URL }
\expandafter\ifx\csname urlstyle\endcsname\relax
  \providecommand{\doi}[1]{doi:\discretionary{}{}{}#1}\else
  \providecommand{\doi}{doi:\discretionary{}{}{}\begingroup
  \urlstyle{rm}\Url}\fi

\bibitem{Lee2011}
\textsc{D.~S. Lee}, \textsc{L.~F. Gonzalez}, \textsc{J.~Periaux}, and
  \textsc{K.~Srinivas}, \enquote{{Efficient hybrid-game strategies coupled to
  evolutionary algorithms for robust multidisciplinary design optimization in
  aerospace engineering},} \emph{IEEE Transactions on Evolutionary
  Computation}, \textbf{15}, \emph{2}, 133 (2011); {10.1109/TEVC.2010.2043364}.

\bibitem{Yang2010b}
\textsc{X.-S. Yang}, \emph{{Engineering Optimization}}, John Wiley {\&} Sons,
  Hoboken (2010).

\bibitem{Moller2001}
\textsc{P.~M{\"{o}}ller}, \textsc{D.~G. Madland}, \textsc{A.~J. Sierk}, and
  \textsc{A.~Iwamoto}, \enquote{{Nuclear Fission Modes and Fragment Mass
  Asymmetries in a Five-dimensional Deformation Space.}} \emph{Nature},
  \textbf{409}, \emph{6822}, 785 (2001); {10.1038/35057204}.

\bibitem{Do2011}
\textsc{J.-M. Do}, \textsc{J.-J. Lautard}, \textsc{A.-M. Baudron},
  \textsc{S.~Douce}, and \textsc{G.~Arnaud}, \enquote{{Use of Meta-Heuristics
  for Design of Fuel Loading Pattern in Light Water Reactors Comprising Some
  Radial and Axial Heterogeneities},} \emph{2011 IEEE International Parallel
  and Distributed Processing Symposium}, 374--380 (2011);
  {10.1109/IPDPS.2011.175}.

\bibitem{Guler2010}
\textsc{O.~Guler}, \emph{{Foundations of Optimization}}, Springer, New York
  (2010).

\bibitem{Storn1997}
\textsc{R.~Storn} and \textsc{K.~Price}, \enquote{{Differential Evolution - A
  Simple and Efficient Heuristic for Global Optimization over Continuous
  Spaces},} \emph{Journal of Global Optimization}, \textbf{11}, 341 (1997);
  {10.1023/A:1008202821328}.

\bibitem{Marichelvam2014}
\textsc{M.~K. Marichelvam}, \textsc{T.~Prabaharan}, and \textsc{X.~S. Yang},
  \enquote{{A discrete firefly algorithm for the multi-objective hybrid
  flowshop scheduling problems},} \emph{Evolutionary Computation, IEEE
  Transactions on}, \textbf{18}, \emph{2}, 301 (2014);
  {10.1109/TEVC.2013.2240304}.

\bibitem{Yang2009}
\textsc{X.-S. Yang} and \textsc{S.~Deb}, \enquote{{Cuckoo Search via Levy
  Flights},} \emph{Nature {\&} Biologically Inspired Computing}, 210--214
  (2009).

\bibitem{Ouaarab2014}
\textsc{A.~Ouaarab}, \textsc{B.~Ahiod}, and \textsc{X.~S. Yang},
  \enquote{{Discrete Cuckoo Search Algorithm for the Travelling Salesman
  Problem},} \emph{Neural Computing and Applications}, \textbf{24}, \emph{7-8},
  1659 (2014); {10.1007/s00521-013-1402-2}.

\bibitem{Yang2014}
\textsc{X.-S. Yang}, \emph{{Nature-Inspired Optimization Algorithms}},
  Elsevier, London (2014).

\bibitem{Hou2016}
\textsc{J.~Hou}, \textsc{S.~Qvist}, \textsc{R.~Kellogg}, and
  \textsc{E.~Greenspan}, \enquote{{3D in-core fuel management optimization for
  breed-and-burn reactors},} \emph{Progress in Nuclear Energy}, \textbf{88},
  \emph{April}, 58 (2016); {10.1016/j.pnucene.2015.12.002}.

\bibitem{Hu2008}
\textsc{H.~Hu} and \textsc{Q.~Wang}, \enquote{{Study on Composit Material for
  Shielding Mixed Neutron and Gama Rays},} \emph{IEEE transactions on science},
  \textbf{55}, \emph{4}, 2376 (2008).

\bibitem{Burer2012}
\textsc{S.~Burer} and \textsc{A.~N. Letchford}, \enquote{{Non-Convex
  Mixed-Integer Nonlinear Pogramming: A Survey},} \emph{Surveys in Operations
  Research and Management Science}, \textbf{17}, \emph{2}, 97 (2012);
  {10.1016/j.sorms.2012.08.001}.

\bibitem{Floudas2009}
\textsc{C.~A. Floudas} and \textsc{C.~E. Gounaris}, \enquote{{A Review of
  Recent Advances in Global Optimization},} \emph{Journal of Global
  Optimization}, \textbf{45}, \emph{1}, 3 (2009); {10.1007/s10898-008-9332-8}.

\bibitem{Bonami2012}
\textsc{P.~Bonami}, \textsc{M.~Kilin{\c{c}}}, and \textsc{J.~Linderoth},
  \enquote{{Algorithms and Software for Convex Mixed Integer Nonlinear
  Programs},} \textsc{J.~Lee} and \textsc{S.~Leyffer} (Editors), \emph{Mixed
  Integer Nonlinear Programming}, 1--39, Springer, New York;
  {10.1007/978-1-4614-1927-3}.

\bibitem{Hemker2008}
\textsc{T.~Hemker}, \enquote{{Derivative Free Surrogate Optimization for
  Mixed-Integer Nonlinear Black Box Problems in Engineering},} PhD Thesis,
  Informatik der Technischen Universitat Darmstadt (2008).

\bibitem{Regis2005}
\textsc{R.~G. Regis} and \textsc{C.~A. Shoemaker}, \enquote{{Constrained global
  optimization of expensive black box functions using radial basis functions},}
  \emph{J. Global Opt}, 31(1), 153--171 (2005).

\bibitem{Egea2014}
\textsc{J.~A. Egea}, \textsc{D.~Henriques}, \textsc{T.~Cokelaer}, \textsc{A.~F.
  Villaverde}, \textsc{A.~MacNamara}, \textsc{D.-P. Danciu}, \textsc{J.~R.
  Banga}, and \textsc{J.~Saez-Rodriguez}, \enquote{{MEIGO: An Open-Source
  Software Suite Based on Metaheuristics for Global Optimization in Systems
  Biology and Bioinformatics},} \emph{BMC Bioinformatics}, \textbf{15}, 136
  (2014); {10.1186/1471-2105-15-136}.

\bibitem{Yiqing2007}
\textsc{L.~Yiqing}, \textsc{Y.~Xigang}, and \textsc{L.~Yongjian}, \enquote{{An
  Improved PSO Algorithm for Solving Non-convex NLP/MINLP Problems with
  Equality Constraints},} \emph{Computers and Chemical Engineering},
  \textbf{31}, \emph{3}, 153 (2007); {10.1016/j.compchemeng.2006.05.016}.

\bibitem{Tao1998}
\textsc{G.~Tao} and \textsc{Z.~Michalewicz}, \enquote{{Iver-over Operator for
  the TSP},} \emph{Parallel Problem Solving from Nature - PPSN V, 5th
  International Conference}, Amsterdam (1998); {DOI: 10.1007/BFb0056922}.

\bibitem{Lones2014}
\textsc{M.~A. Lones}, \enquote{{Metaheuristics in Nature-Inspired Algorithms},}
  \emph{Proceedings of the 2014 Conference on Genetic and Evolutionary
  Computation}, 1419--1422 (2014); {10.1145/2598394.2609841}.

\bibitem{Sorensen2016}
\textsc{K.~Sorensen}, \textsc{M.~Sevaux}, and \textsc{F.~Glover}, \enquote{{A
  History of Metaheuristics},} \emph{Handbook of Heuristics}, Springer;
  {10.1093/brain/122.11.2197-a}.

\bibitem{Glover1986}
\textsc{F.~Glover}, \enquote{{Paths for Integer Programming},} \emph{Computers
  and Operations Research}, \textbf{13}, \emph{5}, 533 (1986);
  {http://dx.doi.org/10.1016/0305-0548(86)90048-1}.

\bibitem{Fister2013}
\textsc{I.~J. Fister}, \textsc{D.~Fister}, and \textsc{I.~Fister}, \enquote{{A
  Comprehensive Review of CS},} \emph{International Journal of Mathermatical
  Modeling and Numerical Optimization}, \textbf{4}, \emph{4}, 387 (2013).

\bibitem{Wolpert1997}
\textsc{D.~H. Wolpert} and \textsc{W.~G. Macready}, \enquote{{No Free Lunch
  Theorems for Optimization},} \emph{IEEE Transactions on Evolutionary
  Computation}, \textbf{1}, \emph{1}, 67 (1997); {10.1109/4235.585893}.

\bibitem{Levy1994}
\textsc{R.~N. Mantegna}, \enquote{{Fast, Accurate Algorithm for Numerical
  Simulation of Levy Stable Stochastic Processes},} \emph{Physical Review E},
  \textbf{49}, \emph{5} (1994).

\bibitem{Zhao2015}
\textsc{K.~Zhao}, \textsc{R.~Jurdak}, \textsc{J.~Liu}, \textsc{D.~Westcott},
  \textsc{B.~Kusy}, \textsc{H.~Parry}, \textsc{P.~Sommer}, and
  \textsc{A.~Mckeown}, \enquote{{Optimal Levy-Flight Foraging in a Finite
  Landscape},} \emph{Journal of the Royal Society Interface}, \textbf{12}
  (2015).

\bibitem{Tran2004}
\textsc{T.~Tran}, \textsc{T.~T. Nguyen}, and \textsc{H.~L. Nguyen},
  \enquote{{Global Optimization using Levy Flights},} \emph{Second National
  Symposium on Research, Development and Application of Information and
  Communication Technology}, Hanoi (2004).

\bibitem{Witze2010}
\textsc{A.~Witze}, \enquote{{Sharks Use Math to Hunt},}
  (2010)\urlprefix\url{https://www.sciencenews.org/article/sharks-use-math-hunt}.

\bibitem{Pantaleo2009}
\textsc{E.~Pantaleo}, \textsc{P.~Facchi}, and \textsc{S.~Pascazio},
  \enquote{{Simulations of L{\'{e}}vy flights},} \emph{Physica Scripta},
  \textbf{T135} (2009).

\bibitem{Viana2013}
\textsc{F.~A.~C. Viana}, \enquote{{Things You Wanted to Know About the Latin
  Hypercube Design and Were Afraid to Ask},} \emph{10th World Congress on
  Structural and Multidisciplinary Optimization}, 1--9, Orlando (2013).

\bibitem{Cioppa2007}
\textsc{T.~M. Cioppa} and \textsc{T.~W. Lucas}, \enquote{{Efficient Nearly
  Orthogonal and Space-Filling Latin Hypercubes},} \emph{Technometrics},
  \textbf{49}, \emph{1}, 45 (2007); {10.1198/004017006000000453}.

\bibitem{Rainville2012}
\textsc{F.-M.~D. Rainville}, \textsc{C.~Gagn{\'{e}}}, \textsc{O.~Teytaud}, and
  \textsc{D.~Laurendeau}, \enquote{{Evolutionary Optimization of
  Low-Discrepancy Sequences},} \emph{ACM Transactions on Modeling and Computer
  Simulation}, \textbf{22}, 1 (2012); {10.1145/2133390.2133393}.

\bibitem{Lin1973}
\textsc{S.~Lin} and \textsc{W.~Kernighan}, \enquote{{An Effective Heurisic
  Algorithm for the Traveling-Salesman Problem},} \emph{Operations Research},
  \textbf{21}, \emph{2}, 498 (1973).

\bibitem{Zhou2014}
\textsc{Y.~Zhou}, \textsc{X.~Ouyang}, and \textsc{J.~Xie}, \enquote{{A Discrete
  Cuckoo Search Algorithm for Travelling Salesman Problem},}
  \emph{International Journal of Collaborative Intelligence}, \textbf{1},
  \emph{1}, 68 (2014).

\bibitem{Hastings1970}
\textsc{W.~K. Hastings}, \enquote{{Monte Carlo Sampling Methods Using Markov
  Chains and Their Applications},} \emph{Biometrika}, \textbf{57}, \emph{1}, 97
  (1970).

\bibitem{Mantegna1994}
\textsc{R.~N. Mantegna} and \textsc{H.~E. Stanley}, \enquote{{Stochastic
  Process with Ultraslow Convergence to a Gaussian: The Truncated Levy
  Flight},} \emph{Physical Review Letters}, \textbf{73}, \emph{22}, 2946
  (1994); {10.1103/PhysRevLett.73.2946}.

\bibitem{Starkweather1991}
\textsc{T.~Starkweather}, \textsc{S.~McDaniel}, \textsc{K.~E. Mathias},
  \textsc{L.~D. Whitley}, and \textsc{C.~Whitley}, \enquote{{A Comparison of
  Genetic Sequencing Operators},} \emph{Proceedings of the fourth International
  Conference on Genetic Algorithms}, 69--76 (1991).

\bibitem{Kennedy1995}
\textsc{J.~Kennedy} and \textsc{R.~Eberhart}, \enquote{{Particle Swarm
  Optimization},} \emph{IEEE International Conference on Neural Networks},
  vol.~4, 1942--1948 (1995); {10.1109/ICNN.1995.488968}.

\bibitem{Walton2011}
\textsc{S.~Walton}, \textsc{O.~Hassan}, \textsc{K.~Morgan}, and \textsc{M.~R.
  Brown}, \enquote{{Modified Cuckoo Search: A New Gradient Free Optimisation
  Algorithm},} \emph{Chaos, Solitons and Fractals}, \textbf{44}, \emph{9}, 710
  (2011); {10.1016/j.chaos.2011.06.004}.

\bibitem{Egea2010}
\textsc{J.~A. Egea}, \textsc{R.~Marti}, and \textsc{J.~R. Banga}, \enquote{{An
  Evolutionary Method for Complex Process Optimization},} \emph{Computers {\&}
  Operations Research}, \textbf{37}, \emph{1}, 315 (2010).

\bibitem{Back1993}
\textsc{T.~B{\"{a}}ck} and \textsc{H.-P. Schwefel}, \enquote{{An Overview of
  Evolutionary Algorithms for Parameter Optimization},} \emph{Evolutionary
  Computation}, \textbf{1}, \emph{1}, 1 (1993); {10.1162/evco.1993.1.1.1}.

\bibitem{Mathworks2015}
\textsc{Mathworks}, \enquote{{MatLab: Global Optimization Toolbox User's Guide
  R2015b},}  (2015).

\bibitem{Ouyang2013}
\textsc{X.~Ouyang}, \textsc{Y.~Zhou}, \textsc{Q.~Luo}, and \textsc{H.~Chen},
  \enquote{{A Novel Discrete Cuckoo Search Algorithm for Spherical Traveling
  Salesman Problem},} \emph{Applied Mathematics and Information Sciences},
  \textbf{7}, \emph{2}, 777 (2013); {10.12785/amis/070248}.

\bibitem{Cagnina2008}
\textsc{L.~C. Cagnina}, \textsc{S.~C. Esquivel}, and \textsc{C.~A. {Coello
  Coello}}, \enquote{{Solving Engineering Optimization Problems with the Simple
  Constrained Particle Swarm Optimizer},} \emph{Informatica (Ljubljana)},
  \textbf{32}, \emph{3}, 319 (2008).

\bibitem{Akhtar2002}
\textsc{S.~Akhtar}, \textsc{K.~Tai}, and \textsc{T.~Ray}, \enquote{{A
  Socio-Behavioural Simulation Model for Engineering Design Optimization},}
  \emph{Engineering Optimization}, \textbf{34}, \emph{4}, 341 (2002);
  {10.1080/03052150212723}.

\bibitem{Lampinen1999}
\textsc{J.~Lampinen} and \textsc{I.~Zelinka}, \enquote{{Mixed
  Integer-Discrete-Continuous Optimization by Differential Evolution},}
  \emph{MENDEL'99}, Brno (1999).

\end{thebibliography}
%



\end{document}